\documentclass[10pt,twocolumn,letterpaper]{article}

\usepackage[pagenumbers]{wacv} 

\usepackage{graphicx}
\usepackage{amsmath}
\usepackage{amssymb}
\usepackage{booktabs}
\usepackage{nicematrix}
\usepackage{bbding} 
\usepackage{dsfont} 

\usepackage{tikz}
\newcommand{\circled}[2][]{\tikz[baseline=(char.base)]{
  \node[shape=circle, draw, inner sep=1pt, font=\sffamily\tiny] (char) {\phantom{\ifblank{#1}{#2}{#1}}};
  \node at (char.center) {\makebox[0pt][c]{\sffamily\tiny #2}};
}}

\usepackage{sg-macros}

\newcommand{\videof}{{\mathbf{f}^\text{V}}}
\newcommand{\videop}{{\mathbf{p}^\text{V}}}
\newcommand{\diagramf}{{\mathbf{f}^\text{I}}}
\newcommand{\diagramp}{{\mathbf{p}^\text{I}}}
\newcommand{\query}{{\mathbf{q}}}

\newcommand{\subsubsubsection}[1]{\noindent\textbf{#1}}

\usepackage[pagebackref,breaklinks,colorlinks]{hyperref}

\usepackage[capitalize]{cleveref}
\crefname{section}{Sec.}{Secs.}
\Crefname{section}{Section}{Sections}
\Crefname{table}{Table}{Tables}
\crefname{table}{Tab.}{Tabs.}

\def\wacvPaperID{126} 
\def\confName{WACV}
\def\confYear{2025}

\begin{document}

\title{Temporally Grounding Instructional Diagrams in Unconstrained Videos}

\author{Jiahao Zhang$^{1}$\quad
Frederic Z. Zhang$^{2}$\quad
Cristian Rodriguez$^{2}$\\
Yizhak Ben-Shabat$^{1,3}$\quad
Anoop Cherian$^{4}$\quad
Stephen Gould$^{1}$\\
$^1$The Australian National University\quad
$^2$The Australian Institute for Machine Learning\\
$^3$Technion Israel Institute of Technology\quad
$^4$Mitsubishi Electric Research Labs\\
{\tt\small $^1$\{first.last\}@anu.edu.au}\quad
{\tt\small $^2$\{first.last\}@adelaide.edu.au}\\
{\tt\small $^3$sitzikbs@gmail.com}\quad
{\tt\small $^4$cherian@merl.com}
}
\maketitle

\begin{figure*}[tb]
  \centering
  \includegraphics[width=\textwidth]{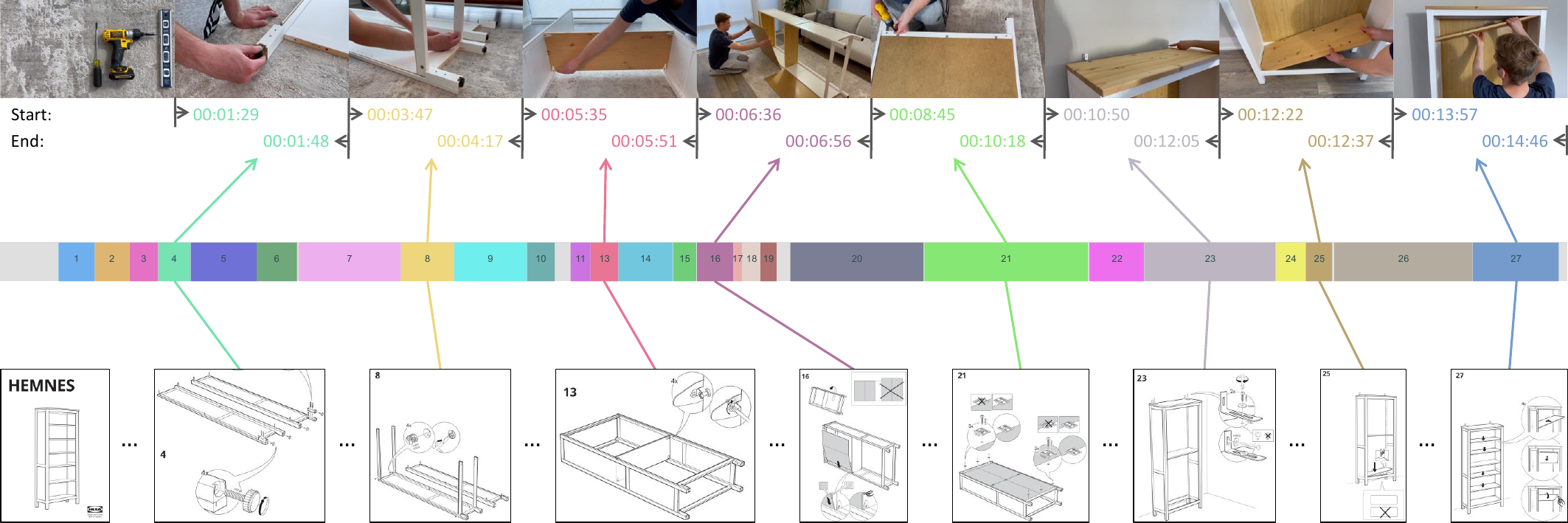}
  \caption{An illustration of the temporal instructional diagram grounding task between a YouTube video (top) \href{https://www.youtube.com/watch?v=xPNkHAii3fU}{xPNkHAii3fU} and an Ikea furniture manual (bottom) \href{https://www.ikea.com/au/en/p/hemnes-bookcase-white-stain-00352894/}{00352894}. This task aims to predict the start and end timestamps for all step diagrams simultaneously.}
  \label{fig:teaser}
  \vspace{-0.2cm}
\end{figure*}

\begin{abstract}
  We study the challenging problem of simultaneously localizing a sequence of queries in the form of instructional diagrams in a video. This requires understanding not only the individual queries but also their interrelationships. However, most existing methods focus on grounding one query at a time, ignoring the inherent structures among queries such as the general mutual exclusiveness and the temporal order. Consequently, the predicted timespans of different step diagrams may overlap considerably or violate the temporal order, thus harming the accuracy. In this paper, we tackle this issue by simultaneously grounding a sequence of step diagrams. Specifically, we propose composite queries, constructed by exhaustively pairing up the visual content features of the step diagrams and a fixed number of learnable positional embeddings. Our insight is that self-attention among composite queries carrying different content features suppress each other to reduce timespan overlaps in predictions, while the cross-attention corrects the temporal misalignment via content and position joint guidance. We demonstrate the effectiveness of our approach on the IAW dataset for grounding step diagrams and the YouCook2 benchmark for grounding natural language queries, significantly outperforming existing methods while simultaneously grounding multiple queries.
\end{abstract}

\section{Introduction}
\label{sec:intro}

Instructional content has become increasingly prevalent, spanning a wide array of applications from DIY projects to educational tutorials. 
One area where this is particularly evident is in instructional videos, which have surged in popularity on platforms like YouTube. 
These videos often provide step-by-step guides on a variety of tasks, including assembly~\cite{ben2021ikea, zhang2023aligning}, cooking~\cite{rohrbach2012script, regneri2013grounding, zhou2018towards, damen2022rescaling}, applying makeup~\cite{wang2019youmakeup}, how-to~\cite{miech2019howto100m} topics and general tasks~\cite{tang2019coin}. 
These guides often boil down to a simple breakdown of complex procedures into manageable and often sequential steps, enabling viewers to easily follow and replicate the processes. 
Nevertheless, it could sometimes be difficult to navigate these lengthy videos---some lasting longer than an hour!---if one intends to find a specific instruction or step.

Among the above-mentioned datasets, the recently proposed Ikea Assembly in the Wild (IAW) dataset~\cite{zhang2023aligning} offers a challenging and real-world setting for research.
The IAW dataset distinct itself from others in two main aspects. 
First, the IAW provides temporal correspondences between segments and step diagrams from instruction manual book, rather than textual descriptions.
Second, there are inherent structures among these step-by-step diagrams within a manual book, where the corresponding timespans are naturally expected to be mutually exclusive and sequentially ordered. 
However, the videos are obtained from YouTube and annotated via crowd-sourcing such that there is no guarantee that the video creators would strictly follow the instruction manual during assembly like those under lab environment.
Additionally, corresponding video segments can overlap with each other considerably (\eg, multiple assemblers, two steps concurrently), and some steps might have multiple segments (\eg, repeating two steps alternately) or none (\eg, optional steps). 
Therefore, the structural prior between queries can only be served as weak general guidance rather than strict constraints.
Consequently, we formulate the task as predicting timestamps for each step diagram from a publicly accessible manual, given an unconstrained video of assembling a piece of furniture, as shown in~\cref{fig:teaser}.

In this work, we propose a novel approach to simultaneously ground a sequence of instructional step diagrams in a video.
Drawing insights from DETR-based temporal grounding methods~\cite{lei2021detecting,jang2023knowing}, we argue that given the interdependencies among the steps in the instruction diagrams of a manual, it may be beneficial to jointly localize all the steps to the respective segments in the associated video rather than localizing one step at a time.
To ground all step diagrams at the same time, we first directly concatenate feature representation of one step diagram (content query) with one learnable positional query (position query) to create one composite query.
Then, the final composite queries, which fed into the decoder, are constructed by exhaustively pairing up all possible combinations of content and position queries.
Such that, these composite queries are able to interact with each other via self-attention, allowing the modeling of the inherent structures of a step diagram sequence.
Additionally, since each composite query already carries the respective step diagram, the correspondence is known, eliminating the need for applying classification~\cite{lei2021detecting} or matching~\cite{woo2022explore} at the final output layer.
It is also worth noting that the position queries of composite queries act as input-agnostic learned temporal bias, similar to templates~\cite{carion2020end, meng2021conditional} in object detection transformers, which guide the cross-attention temporally in conjunction with the semantics of the video and diagram.

Our primary contribution is a novel paradigm for a detection transformer model capable of temporally grounding a sequence of step diagrams simultaneously. 
Building on Moment-DETR~\cite{lei2021detecting}, we demonstrate that by incorporating the inherent structure of a sequence of step diagrams via constructing composite queries, the model can reduce the overlapping of predicted timespans and enhance the temporal correlation with respect to the ground truth. 
Visualizing the cross-attention maps reveals that the content and position joint cross-attention enables the model to do temporal corrections. 
To tailor the transformer decoder module to our problem, we design various attention masks in the self-attention module and experiment with different choices in the transformer decoder. 
Our method is complementary to those focusing on improving the performances for singleton queries. 
By integrating our method into a stronger single-query model, EaTR~\cite{jang2023knowing}, we observed a consistent performance boost. 
Our method significantly outperforms existing ones on the IAW~\cite{zhang2023aligning} dataset for step diagram sequence grounding and on YouCook2~\cite{shimomoto2022towards} benchmark for sentence sequence grounding, demonstrating its effectiveness in simultaneously grounding multiple queries.

\section{Related Work}
\label{sec:work}

\subsubsubsection{Temporal sentence grounding in videos (TSGV)} aims to locate a segment in a video given a textual query. 
There has been significant research interests~\cite{zhang2023temporal} since its introduction by Hendricks \etal~\cite{anne2017localizing} and Bao \etal~\cite{gao2017tall}. 
Research in this area primarily divides into two approaches: proposal-based and proposal-free methods. 
Proposal-based methods adapts a \textit{propose-then-match} strategy, initially using a sliding window technique~\cite{gao2017tall, anne2017localizing} to create candidate proposals. 
To improve efficiency and reduce redundancy, anchor-based methods are developed~\cite{chen2018temporally, zhang2019man, yuan2019semantic}.
A notable contribution in this category is the 2D-TAN model~\cite{zhang2020learning}, which utilizes a 2D map to enumerate all possible anchor proposals and consider their temporal relationships. 
Later, proposal-free methods aim to identify the target moment directly without generating proposals. 
This includes regression-based methods~\cite{yuan2019find, ghosh2019excl}, that directly regress the start and end time of the target moment, and span-based methods~\cite{zhang2020span, zhang2021parallel, rodriguez2020proposal, rodriguez2021locformer, rodriguez2021dori, rodriguez-opazo-etal-2023-memory} that calculate the probability of each video frame or clip being the start or end of the target.

\begin{figure*}[tb]
  \centering
  \includegraphics[width=.85\textwidth]{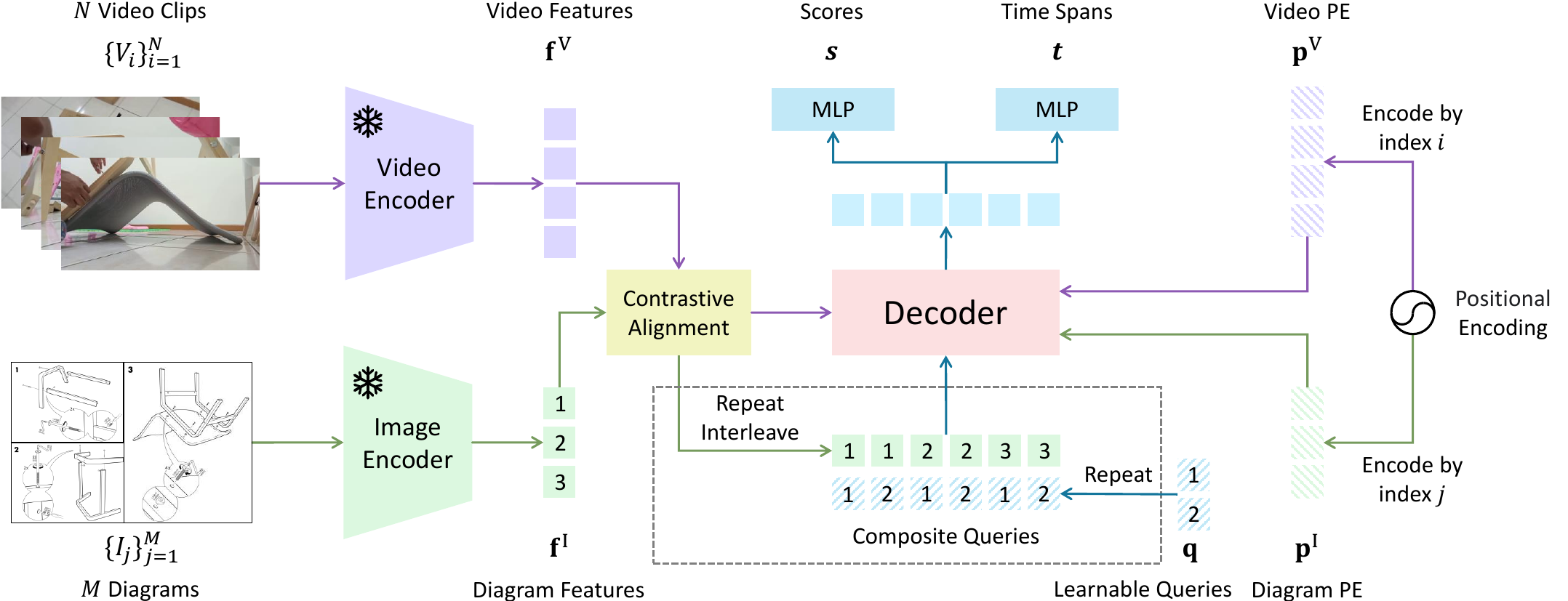}
  \caption{Schematic of our overall pipeline. First, we extract features from the inputs using pre-trained video and image encoders and employ two MLPs to project these features into a unified $D$-dimensional space, resulting in $\videof\in\mathbb{R}^{N\times D}$ and $\diagramf\in\mathbb{R}^{M\times D}$, respectively. Following Zhang \etal~\cite{zhang2023aligning}, first apply contrastive alignment then project these two features into a single joint embedding space. Second, positional encoding (PE) is used to obtain a video $\videop$ and a diagram $\diagramp$ PE vectors. Third, we build composite queries by repeating diagram features and learnable queries. Last, all features, positional encoding, and composite queries are fed into a variant of a transformer decoder, which is specifically designed for this task (\cref{sec:main-method}). The decoder's output is then processed by another two MLPs, tasked with generating scores $\bs$ and timespans $\bt$, respectively.}
  \label{fig:pipeline}
  \vspace{-0.2cm}
\end{figure*}

\subsubsubsection{DETR-based TSGV methods} draw significant inspiration from the success of DETR~\cite{carion2020end}, adapting its concepts to the regression-based, proposal-free domain of TSGV. 
By conceptualizing the grounding task as a 1D variant of object detection, the pioneering work Moment-DETR~\cite{lei2021detecting} first adapted DETR and incorporated saliency for each sentence as guidance.
Subsequent works have built on this foundation with various enhancements: 
QD-DETR~\cite{moon2023query} improved predictions by detecting not only positive but also negative queries; 
MS-DETR~\cite{wang2023ms} merged the concepts of 2DTAN and DETR, introducing sampling moment-moment interactions,
EaTR~\cite{jang2023knowing} learned pseudo events from video as priors injected into queries, 
UniVTG~\cite{lin2023univtg}, TaskWeave~\cite{yang2024task}, and UVCOM~\cite{xiao2024bridging} studied the joint learning of moment retrieval and highlight detection.
CG-DETR~\cite{moon2023correlation} guided the cross-attention via word-level textual conditions to reduce wrong correspondences. 
RGTR~\cite{sun2024diversifying} aimed to make the predicted region of each query more distinct and concentrated.
However, all the above methods are designed for the single-query scenario. 
The exception, LVTR~\cite{woo2022explore}, specifically addresses the challenge of multi-sentence grounding by concatenating a fixed four sentences with video features before encoding, however performs poorly when the number of queries varies across the dataset like in the IAW.
Where we tackle this issue by constructing composite queries.

\subsubsubsection{Dense events grounding in videos} was recently introduced by Bao \etal~\cite{bao2021dense}. 
This task extends TSGV tasks by aiming to ground all sentences simultaneously. 
Bao \etal also introduced a model titled DepNet that adopts the \textit{propose-then-match} approach. 
Another study by Tan \etal~\cite{tan2023hierarchical} explores the hierarchical structure of words, sentences, and paragraphs with HSCNet. 
Jiang \etal~\cite{jiang2022semi} proposed SVPTR, advancing the task into the semi-supervised learning domain. 
The work most similar to ours is PRVG~\cite{shi2021end}, but differs in two key ways: 1) PRVG eliminates the use of learnable queries, restricting the model to make only one prediction per text query; 2) it fuses video and text features before the decoding process using two encoders. 
All mentioned methods are evaluated on the ActivityNet Captions~\cite{Krishna_2017_ICCV} and TACoS~\cite{regneri2013grounding} datasets. 
In these two datasets, the sentences from ground truth and video segments correspond one-to-one, which means that there is no instance where the same sentence corresponds to multiple timespans, or cases where the one sentence does not have a corresponding timespan. 
This is in contrast to the IAW~\cite{zhang2023aligning} dataset we primarily use.

\section{Temporal Diagram Grounding in Videos}
\label{sec:method}

\subsection{Problem Definition}

Temporal instructional diagram grounding in unconstrained videos is an extension of TSGV task. Instead of grounding a single descriptive sentence, we aim to simultaneously ground a sequence of diagrams extracted from an instruction manual. Formally, given an untrimmed video that has been evenly divided into $N$ clips for feature extraction, we denote these clips by $\{V_i\}_{i=1}^{N}$, and the set of $M$ instructional diagrams corresponding to the video by $\{I_j\}_{j=1}^{M}$, our objective is to develop a model capable of accurately predicting the timespan of each diagram $\bt = (t_s, t_e)$, where $t_s$ and $t_e$ are the normalized start and end time of a segment.

\subsection{Proposed Method}
\label{sec:main-method}

We propose a DETR-based method as illustrated in~\cref{fig:pipeline}. Below is a detailed introduction to the decoder.

\begin{figure*}[tb]
  \centering
  \begin{subfigure}{0.32\linewidth}
    \includegraphics[width=\linewidth]{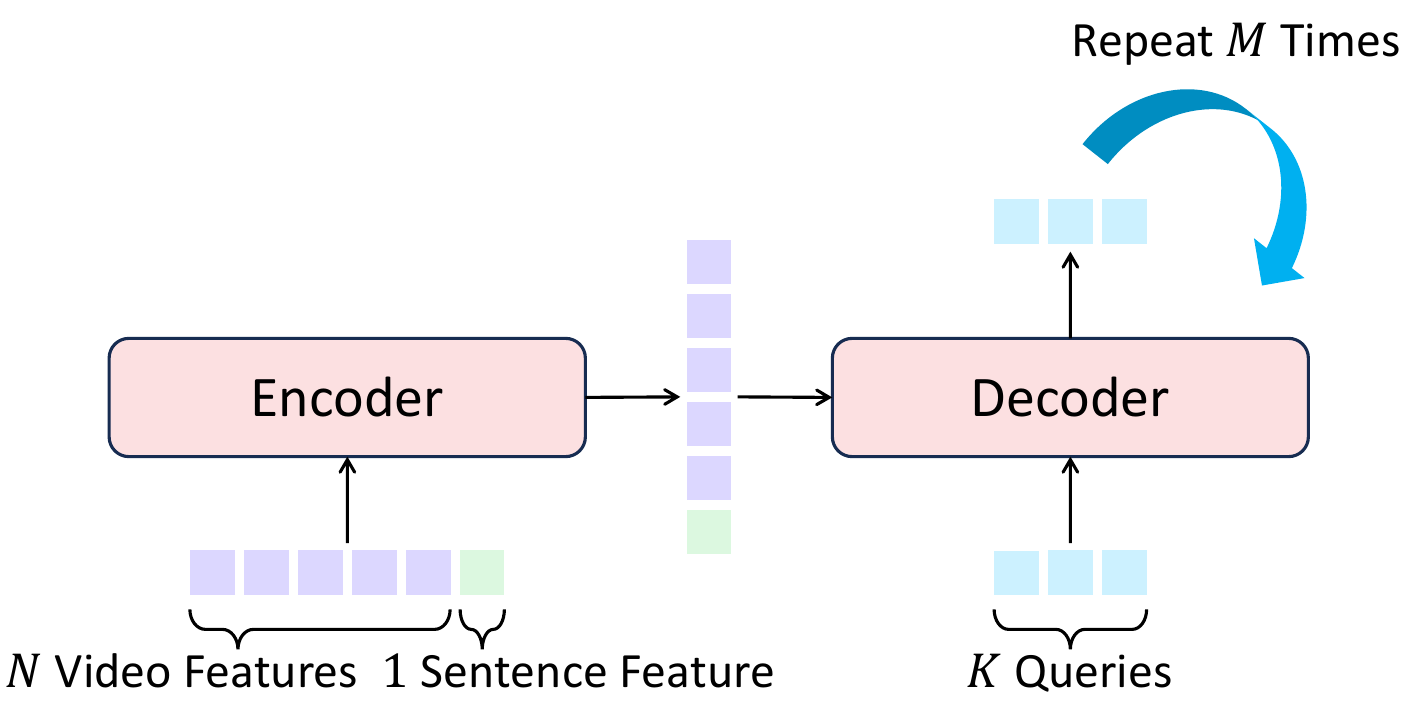}
    \caption{Moment-DETR~\cite{lei2021detecting}}
    \label{fig:moment-detr}
  \end{subfigure}
  \hfill
  \begin{subfigure}{0.32\linewidth}
    \includegraphics[width=\linewidth]{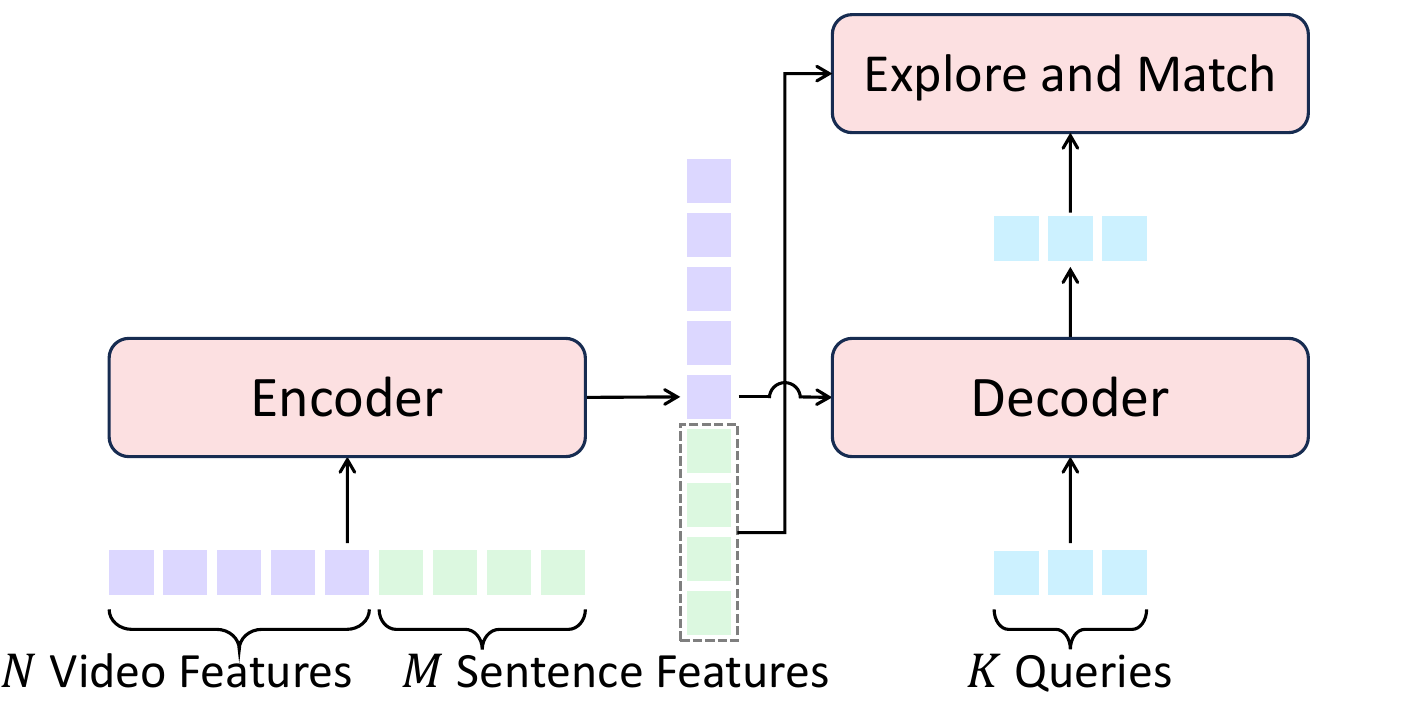}
    \caption{LVTR~\cite{woo2022explore}}
    \label{fig:lvtr}
  \end{subfigure}
  \hfill
  \begin{subfigure}{0.32\linewidth}
    \includegraphics[width=\linewidth]{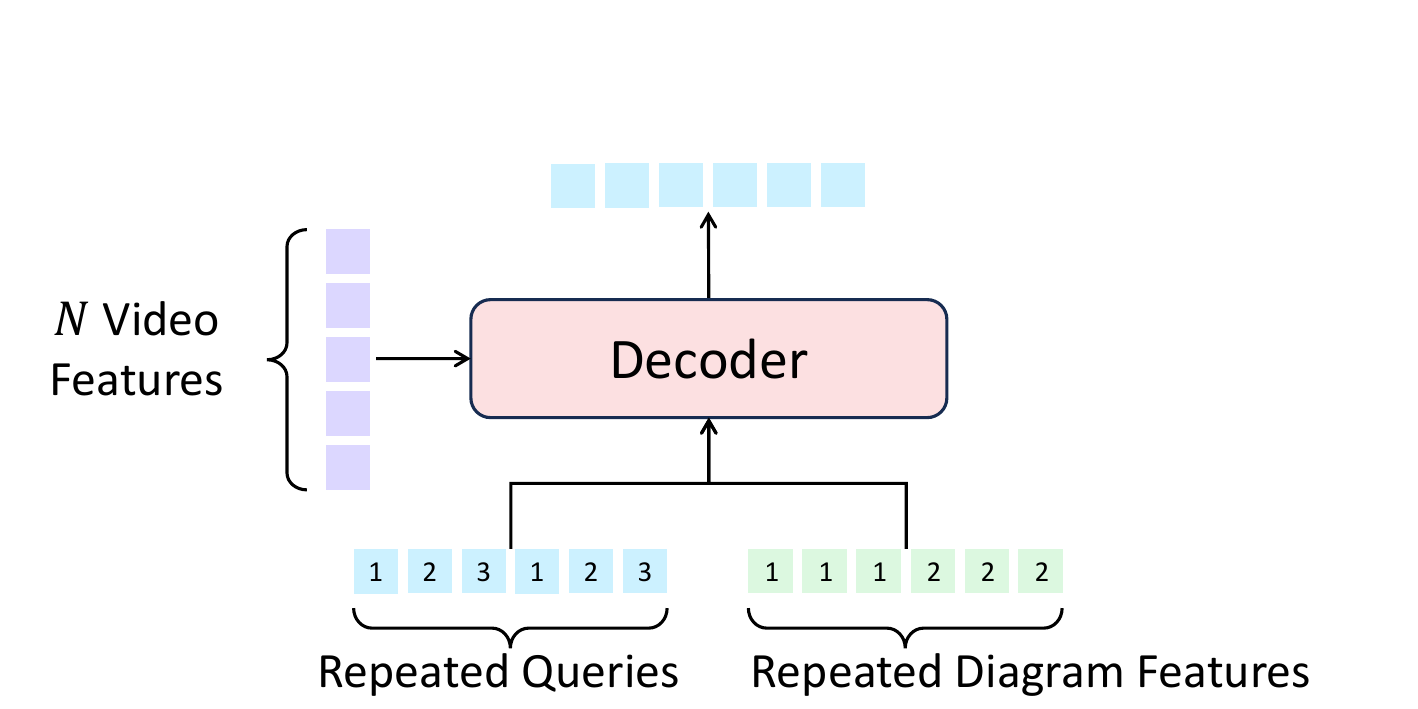}
    \caption{Ours}
    \label{fig:our-all}
  \end{subfigure}
  \caption{Illustration of existing DETR-based temporal sequence grounding models for comparing with our method. We use three learnable queries and two diagrams as an example ($K=3, M=2$). Note the repeat strategies for learnable queries and diagram features in (d) are different, so that they can be intersecting fused.}
  \label{fig:compare}
  \vspace{-0.2cm}
\end{figure*}

\subsubsubsection{Decoder-Based Architecture.}
As depicted in~\cref{fig:compare}, inspired by the DETR~\cite{carion2020end} architecture for object detection, models for TSGV, such as Moment-DETR~\cite{lei2021detecting} (\cref{fig:moment-detr}), typically feature with one encoder for fusing multimodal inputs and one decoder dedicated to processing learnable queries and predicting the timespans. However, for the task of temporal action localization, Liu \etal~\cite{liu2022end} showed through ablation studies that the presence of the encoder may harm the performance. Kim \etal~\cite{kim2023self} delved deeper into this issue, attributing the negative effect to what they termed \textit{temporal collapse}, where the self-attention only focus on a few key elements. Motivated by these findings, we propose to use the diagram features as content priors for the queries and values, video features as keys to the decoder, and remove the encoder, as depicted in~\cref{fig:our-all}.

\subsubsubsection{Composite Query via Duplication.}
\label{sec:composite}
As shown in~\cref{fig:lvtr}, LVTR~\cite{woo2022explore}, albeit being able to ground multiple sentences simultaneously, employs a fixed number of learnable queries, each of which predicts the timespan for one sentence. 
To accommodate a varying sequence length, it has to use a large number ($\ge80$ in IAW) of learnable queries, which results in most of such queries not being matched to any ground truth instances during training, thus not receiving enough gradients to be trained effectively. 
We address this issue by using the feature representations of diagrams as content priors, and only employ a relative small number ($\le10$ in IAW) of learnable queries. 
Each diagram will then be exhaustively paired up with the learnable queries, which can be easily implemented via duplication as shown in the dashed area in~\cref{fig:pipeline}. 
Formally, given $M$ diagram features $\diagramf=\{\diagramf_i\}_{i=1}^M$ and $K$ learnable positional embeddings $\query=\{\query_j\}_{j=1}^K$, the \emph{composite query} is defined as
\begin{align}
    \{
    (\diagramf_i\oplus\query_j) 
    \mid
    (\diagramf_i,\query_j)\in\diagramf\times\query
    \},
\end{align}
where $\oplus$ denotes the concatenation operation and $\times$ is the Cartesian product.
Drawing intuitions from previous work~\cite{carion2020end,meng2021conditional}, queries in the decoder typically encapsulate two key aspects: semantic and positional information. 
As Meng \etal~\cite{meng2021conditional} pointed out, the learnable queries, which are implemented as learnable positional embeddings, serve as positional priors, while the subsequent cross-attention layers inject the relevant semantics into these queries. 
The composite queries have a distinct advantage, in that the semantic information is explicitly encoded into their representations as content priors, before the cross-attention layers.
Consequently, the model is relieved from the challenging task of aggregating semantics, particularly when the number of step diagrams is dynamic, allowing it to concentrate more on the grounding task.
During training, we perform Hungarian matching~\cite{carion2020end} only between the set of composite queries that share the same content prior, \ie, diagram features and the ground truth for that step diagram, significantly simplifying the matching process.

\begin{figure}[tb]
  \centering
  \begin{subfigure}{0.37\linewidth}
    \includegraphics[width=\linewidth]{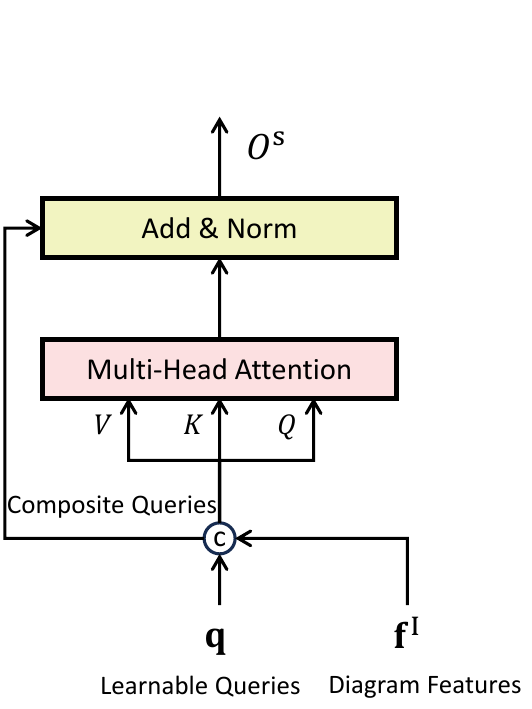}
    \caption{Self-Attention Module}
    \label{fig:decoder-sa}
  \end{subfigure}
  \hfill
  \begin{subfigure}{0.62\linewidth}
    \includegraphics[width=\linewidth]{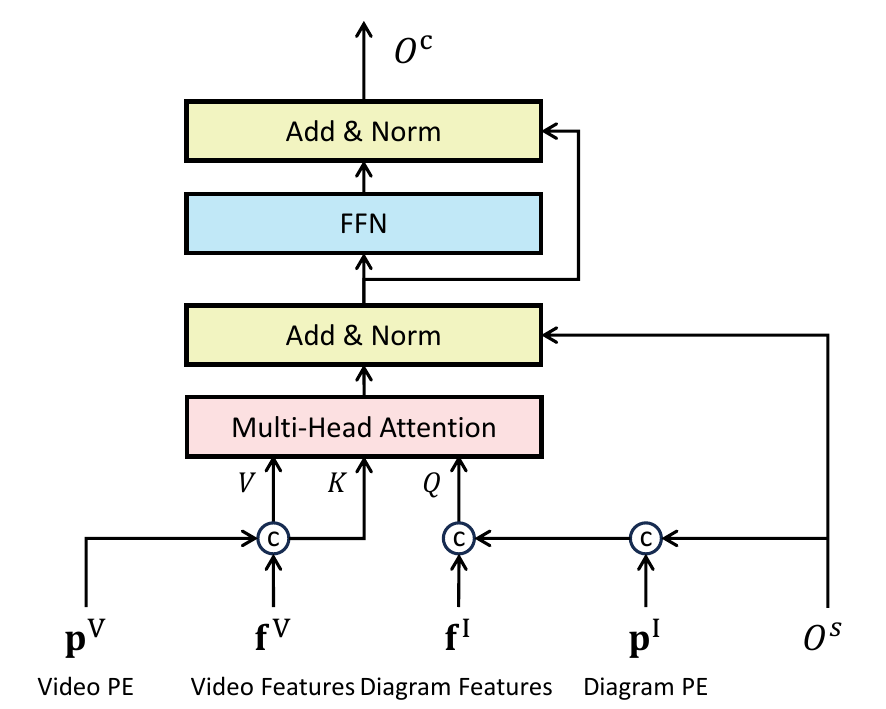}
    \caption{Cross-Attention Module}
    \label{fig:decoder-ca}
  \end{subfigure}
  \caption{Illustration of the decoder architecture. We use \circled{C} to represent first concatenate then project back operation. The concatenation order is always first features then positional encodings.}
  \label{fig:decoder}
  \vspace{-0.4cm}
\end{figure}

\begin{figure*}[tb]
  \centering
  \begin{subfigure}{0.20\linewidth}
    \includegraphics[width=\linewidth]{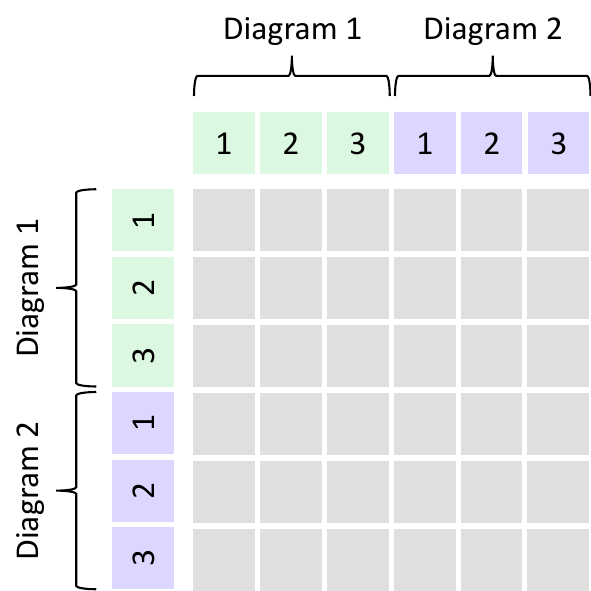}
    \caption{\quad Type A}
    \label{fig:mask-a}
  \end{subfigure}
  \hfill
  \begin{subfigure}{0.20\linewidth}
    \includegraphics[width=\linewidth]{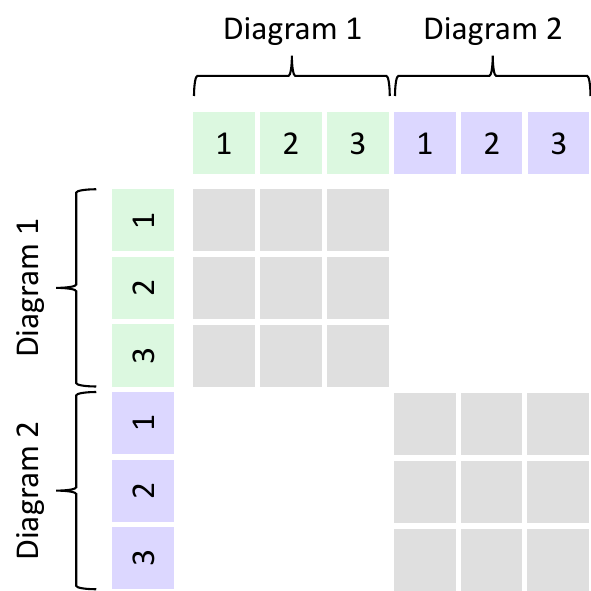}
    \caption{\quad Type B}
    \label{fig:mask-b}
  \end{subfigure}
  \hfill
  \begin{subfigure}{0.20\linewidth}
    \includegraphics[width=\linewidth]{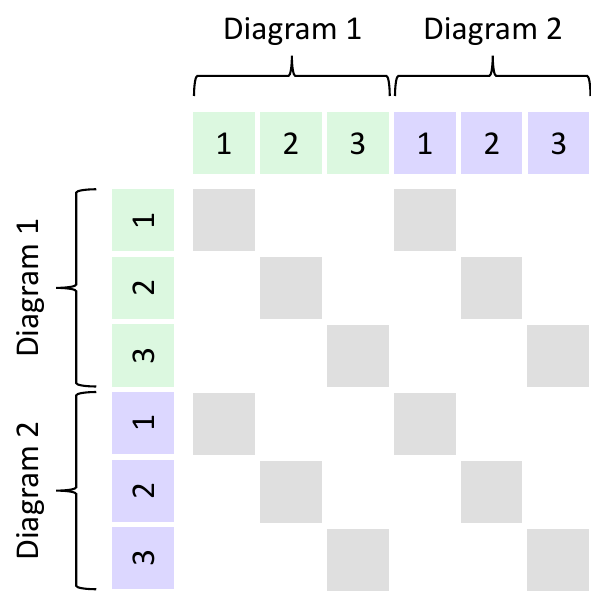}
    \caption{\quad Type C}
    \label{fig:mask-c}
  \end{subfigure}
  \hfill
  \begin{subfigure}{0.20\linewidth}
    \includegraphics[width=\linewidth]{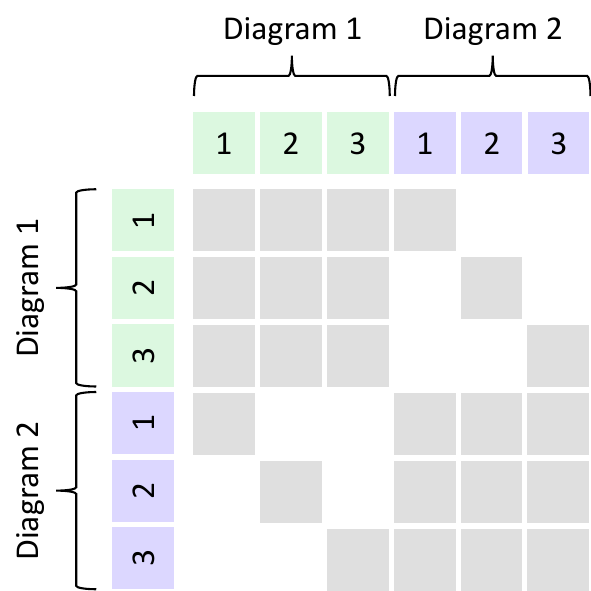}
    \caption{\quad Type D (Ours)}
    \label{fig:mask-d}
  \end{subfigure}
  \caption{Illustration of different types of self-attention mask applied in multi-head attention module in~\cref{fig:decoder-sa}. We use three queries, two diagrams ($K=3,M=2$) in this example. We use white to indicate the attention weights that are masked and ignored.}
  \label{fig:mask}
  \vspace{-0.2cm}
\end{figure*}

\subsubsubsection{Content and Position Jointly Guided Cross-Attention.}
Recall that the attention mechanism in transformers~\cite{vaswani2017attention} uses dot-product between the keys and queries ($QK\transpose$) as a similarity measurement to direct information aggregation. We delve into the intuitions behind this mechanism and tailor the design to our specific problem. Following previous practice~\cite{meng2021conditional}, we denote the keys and queries by $K_c$ and $Q_c$ respectively, as they primarily contain content features, and their positional embeddings by $K_p$ and $Q_p$. In our model, the content aspect is captured by the features extracted from the frozen encoders (\cref{fig:pipeline}), while positional information is embedded using the sinusoidal functions. Drawing inspiration from~\cite{meng2021conditional,zhang2023exploring}, the content features are concatenated to the positional embeddings, instead of being added, as it removes two cross-terms ($Q_c K_p\transpose,Q_p K_c\transpose$) that may be noisy. Thus, the attention weights before softmax and scaling becomes
\begin{equation}
  (Q_c\oplus Q_p)(K_c\oplus K_p) = Q_c K_c \transpose + Q_p K_p \transpose,
  \label{eq:a}
\end{equation}
where $Q_c K_c \transpose$ calculates the semantic similarity between video and diagram features while $Q_p K_p \transpose$ measures the similarity between video positional encodings and the learnable positional embeddings. As such, the cross-attention mechanism is guided by both the semantic and positional cues.

\subsubsubsection{Masked Self-Attention.}
\label{sec:mask}
The self-attention module among queries in each layer of the decoder, as shown in~\cref{fig:decoder-sa}, has been shown to exhibit suppressive behaviors~\cite{carion2020end, zhang2022efficient}, which eliminate the needs of non maximum suppression (NMS) on predictions. A composite query by design could suppress other competing composite queries to secure the highest score at the end. But with the duplication mechanism mentioned above, the vanilla self-attention (type A in~\cref{fig:mask-a}) may introduce some undesired pairs of suppression. Specifically, for a given diagram associated with $K$ queries, the intended behavior is for one query to effectively suppress the others, mirroring the block diagonal pattern seen in type B (\cref{fig:mask-b}). Additionally, when the same query is used across $M$ different diagrams, it is necessary for one diagram to suppress the others to reduce overlaps, as depicted by the sub-diagonals in type C (\cref{fig:mask-c}). However, attention scores between different queries and diagrams might inadvertently decrease the model's adaptability by suppressing all potential timespan prediction pairs. To mitigate this, we implement the self-attention mask type D (\cref{fig:mask-d}) to filter out these undesirable pairings.

\subsubsubsection{Diagram Order Prior.}
We incorporate the positional encoding of the diagram indices $\diagramp$, into the output from the self-attention module, as depicted in~\cref{fig:decoder-ca}. This is to ensure coherence with the positional encoding of the video $\videop$, effectively aligning both modalities in terms of their temporal ordering information.

\subsubsubsection{Choice of the Attention Value Feature.}
\label{sec:value}
The core learning objective of the decoder is to derive a mapping function that translates the input $Q,K,V$ into some outputs that implicitly represent scores $\bs$ and timespans $\bt$. The outputs are essentially weighted aggregates of the values $V$, gathering from both self-attention $V^s$ and cross-attention~$V^c$. For the timespan predictions, the mapping can be learned either explicitly from the positional encoding space $(V\subseteq\{\videop,\diagramp,\query\})$ or implicitly from the feature space $(V\subseteq\{\videof,\diagramf\})$. To maximize adaptability and leverage information from both these spaces, we opt to concatenate these two then project them back to their original dimensionality as shown in~\cref{fig:decoder}, thereby affording the model greater flexibility in integrating diverse cues.

\subsection{Losses}

To train our model, we adopt the moment retrieval loss from Moment-DETR~\cite{lei2021detecting}.
Formally, denote the predicted timespan for a composite query by $\hat{\bt}_{i,j} \in [0,1]^2$,
where $ i \in \{1, \ldots, M\}$ indexes a step diagram and $ j \in \{1, \ldots, K\}$ indexes a learnable positional embedding.
Denote the normalized prediction scores by $\hat{\bs}_{i, j} \in \reals$.
Following previous practice,
we pad the ground truth with $\varnothing$ to size $MK$ and use the Hungarian algorithm~\cite{kuhn1955hungarian} for bipartite matching.
For brevity of exposition,
we use $\bt^{\star}_{i,j}$ and $\bs^{\star}_{i,j}$ to denote the ground truth for composite query $(i,j)$,
where $\bs^{\star}_{i,j}=0$ if the matched ground truth is an empty set and otherwise $\bs^{\star}_{i,j}=1$ and there are $C_i$ out of $K$ matched for $i$-th diagram.
We then employ a combination of three discrepancy measurements: a $L_1$ loss and a generalized IoU loss~\cite{rezatofighi2019generalized} between predicted and ground truth timespans,
\begin{align}
    \mathcal{L}_{L_1}&=\frac{1}{M}\sum_{i=1}^{M}\frac{1}{C_i}\sum_{j=1}^K \mathds{1}_{\{\bs^{\star}_{i,j}=1\}} \lVert \hat{\bt}_{i,j}-\bt_{i,j}^\star \rVert_1, \\
    \mathcal{L}_{\text{gIoU}}&=\frac{1}{M}\sum_{i=1}^{M}\frac{1}{C_i}\sum_{j=1}^{K} \mathds{1}_{\{\bs^{\star}_{i,j}=1\}} gIoU(\hat{\bt}_{i,j},\bt_{i,j}^\star);
\end{align}
and a cross entropy loss to classify each predicted timespan being foreground or background, depending on whether it is matched by Hungarian algorithm,
\begin{align}
    \mathcal{L}_{\text{score}}=-\frac{1}{MK}\sum_{i=1}^{M}\sum_{j=1}^{K} \mathbf{s}_{i,j}^\star\log(\hat{\bs}_{i,j}).
\end{align}
Thus, the overall target is
\begin{align}
    \mathcal{L}=\lambda_{L_1}\mathcal{L}_{L_1}+\lambda_{\text{gIoU}}\mathcal{L}_{\text{gIoU}}+\lambda_{\text{score}}\mathcal{L}_{\text{score}},
\end{align}
where $\lambda_{L_1}, \lambda_{\text{gIoU}}, \lambda_{\text{score}}$ are hyperparameters used for balancing each loss term.

\section{Experiments and Results}
\label{sec:exp}

\subsection{Experiment Setup}

\subsubsubsection{Datasets.}
The primary dataset used for designing and evaluating our model is the Ikea Assembly in the Wild (IAW)~\cite{zhang2023exploring}, which features 1,007 untrimmed assembly videos obtained from YouTube across 420 different furniture, each with corresponding instructional step diagrams. The IAW dataset is divided into training, validation, and testing splits with 9,859, 2,210, and 3,512 timespans, respectively. Additionally, we also evaluate our model on YouCook2~\cite{zhou2018towards}, which has 2000 untrimmed videos from 89 cooking recipes. The procedure steps for each video are annotated with temporal boundaries and described by imperative English sentences. YouCook2 has 10,337 and 3,492 ground truth timespans for train and validate, respectively.

\subsubsubsection{Preprocessing.}
For the IAW dataset, we leverage the recent VideoMAEv2~\cite{wang2023videomae} and DINOv2~\cite{oquab2023dinov2} to extract features from videos and diagrams, respectively. Then, following Zhang \etal~\cite{zhang2023aligning}, contrastive learning is applied at the feature level to enhance the representational alignment (refer to supplementary for details). For the YouCook2 dataset, I3D~\cite{carreira2017quo} video and BERT~\cite{devlin2018bert} text features (averaged word features as a sentence feature) are extracted and established benchmarks~\cite{rodriguez2020proposal, rodriguez2021dori, shimomoto2022towards} are used to ensure a fair and consistent comparison.

\subsubsubsection{Sliding Window Sampler.}
A sliding window sampler is implemented to temporally sub-samples the videos into chunks with given window size $w$ and step stride $s$. Furthermore, by aggregating sampled segments from various sliding window configurations (varying both sizes and strides) we enhance the diversity of our training data.

\subsubsubsection{Training and Inference.}
In line with the approach used by 2DTAN~\cite{zhang2020learning}, we normalize the length of video features to 256 through interpolation. For the sampling process, we configure the window sizes $w \in \{128, 256, 512, 1024, \infty\}$ and strides $s \in \{32, 64, 128, 256, \infty\}$, with $\infty$ indicating the use of the entire video. The number of queries is set to 3 based on empirical findings. We use AdamW~\cite{loshchilov2017decoupled} as the optimizer with a multi-step learning rate scheduler starting at $10^{-4}$ and a weight decay of $10^{-4}$. Training is conducted on a single A100 GPU for 60 epochs, using a batch size of 16, which roughly takes 20 hours. The model demonstrating the best performance on the validation split is selected for reporting results on the test split of the IAW dataset. For YouCook2, since they only provide ground truth for validation split, we use that for both model selection and result report. The code will be made available upon acceptance.

\subsubsubsection{Evaluation Metrics.}
To evaluate our model, we adapt standard recall at $k$ given a temporal intersection over union threshold $m$ ($\text{R}@k,\text{IoU}=m$) and the mean intersection over union (mIoU) metrics. Conventionally, we choose $k=1$ and $m \in \{0.3,0.5,0.7\}$.

\subsection{Main Results}

\begin{table}[tb]
    \centering
    \scriptsize
    \newcolumntype{Z}{>{\centering\arraybackslash}X}
    \caption{Evaluation results of different methods on IAW~\cite{zhang2023aligning} dataset test split. Mode \textit{One} denotes one diagram at a time, and \textit{All} means all diagrams at the same time by a single run. All the models listed are trained with the same aligned features unless specified otherwise. The values in bold represent the best results, while underlined indicate the second.}
    \begin{NiceTabularX}{\linewidth}{
            @{} l | Z | *{3}{Z} | Z
        }
        \toprule
        \Block{2-1}{Method}
        & \Block{2-1}{Mode}
        & \Block{1-3}{R@1, IoU=}
        &
        &
        & \Block{2-1}{mIoU}
        \\
        \cmidrule(lr){3-5}
        &
        & 0.3
        & 0.5
        & 0.7
        \\
        \midrule
        Random
        & -
        & 1.809
        & 0.254
        & 0.057
        & 4.801
        \\
        \midrule
        2D-TAN~\cite{zhang2020learning} conv
        & \Block{4-1}{One}
        & 31.24
        & 18.94
        & 8.030
        & 20.51
        \\
        2D-TAN~\cite{zhang2020learning} pool
        &
        & 32.94
        & 20.02
        & 8.170
        & 21.21
        \\
        Moment DETR~\cite{lei2021detecting}
        & 
        & 34.00
        & 18.34
        & 7.290
        & 16.60
        \\
        EaTR~\cite{jang2023knowing}
        & 
        & \underline{38.48}
        & \underline{22.77}
        & \underline{9.540}
        & \underline{24.75}
        \\
        \midrule
        LVTR~\cite{woo2022explore}
        & \Block{3-1}{All}
        & 11.26
        & 4.591
        & 1.112
        & 7.515
        \\
        Ours (Moment DETR)
        & 
        & 37.79
        & 22.74
        & 9.140
        & 23.86
        \\
        Ours (EaTR)
        & 
        & \textbf{42.02}
        & \textbf{26.45}
        & \textbf{11.54}
        & \textbf{27.27}
        \\
        \bottomrule
    \end{NiceTabularX}
    \label{tab:main_result}
    \vspace{-0.2cm}
\end{table}

\begin{table}[tb]
    \centering
    \scriptsize
    \newcolumntype{Z}{>{\centering\arraybackslash}X}
    \caption{Evaluation results on the val split of YouCook2~\cite{zhou2018towards}.}
    \begin{NiceTabularX}{\linewidth}{
            @{} l | Z | *{3}{c} | c
        }
        \toprule
        \Block{2-1}{Method}
        & \Block{2-1}{Text}
        & \Block{1-3}{R@1, IoU=}
        &
        &
        & \Block{2-1}{mIoU}
        \\
        \cmidrule(lr){3-5}
        &
        & 0.3
        & 0.5
        & 0.7
        \\
        \midrule
        DORi~\cite{rodriguez2021dori}
        & -
        & 43.36
        & 30.47
        & 18.24
        & 30.46
        \\
        LocFormer~\cite{rodriguez2021locformer}
        & BERT~\cite{devlin2018bert}
        & \underline{46.76}
        & \underline{31.33}
        & 15.81
        & \underline{30.92}
        \\
        \midrule
        ExCL~\cite{ghosh2019excl}
        & \Block{3-1}{BERT \Snowflake~\cite{shimomoto2022towards}}
        & 26.63
        & 16.15
        & 8.51
        & 18.87
        \\
        TMLGA~\cite{rodriguez2020proposal}
        & 
        & 34.77
        & 23.05
        & 12.49
        & 24.42
        \\
        DORi~\cite{rodriguez2021dori}
        & 
        & 42.27
        & 29.90
        & \underline{18.38}
        & 29.92
        \\
        \midrule
        Ours (EaTR)
        & BERT \Snowflake~\cite{shimomoto2022towards}
        & \textbf{52.95}
        & \textbf{36.28}
        & \textbf{18.50}
        & \textbf{35.32}
        \\
        \bottomrule
    \end{NiceTabularX}
    \label{tab:youcook2}
    \vspace{-0.25cm}
\end{table}

\subsubsubsection{Results on IAW.}
We report the performance of our model on the IAW~\cite{zhang2023aligning} dataset in~\cref{tab:main_result}.
When grounding all diagrams at the same time (mode \textit{All}), LVTR's performance drops significantly. 
It is perhaps due to the challenges posed by its Explore-and-Match (\cref{fig:lvtr}) module, which seeks to align a mismatched number of learnable queries and diagrams. 
The queries in LVTR need to be able to aggregate both the semantic and positional information such that they can be matched to the inputs and also be decoded to timespans.
Where ours already carry semantics.
This is even harder in IAW dataset, since for different videos, the manual books are different. 
Thus, the performance of LVTR is even lower than 2D-TAN and Moment-DETR. 
Built upon Moment-DETR, our multi-query version denoted as ``Ours (Moment DETR)'' further improves performance by a significant margin comparing with its single-query version, showing the importance to incorporate the inherent structure within a sequence of queries, which benefits from the design of composite queries and content position joint guided cross-attention.  
Besides that, we tried to integrate ours techniques to a stronger single-query method EaTR~\cite{jang2023knowing} as ``Ours (EaTR)\footnote{We refer the details of modification to the supplementary.}'', the consistent observation of the performance improvement shows our design could be also beneficial to other DETR-based single-query methods.

\subsubsubsection{Results on YouCook2.}
We report the performance of our model based on EaTR on the YouCook2~\cite{zhou2018towards} dataset in~\cref{tab:youcook2}. 
Notably, unlike our approach, DORi~\cite{rodriguez2021dori} utilizes additional object relationship features. 
Additionally, our sentence features are obtained by averaging pre-extracted word features, whereas LocFormer~\cite{rodriguez2021locformer} and DORi~\cite{rodriguez2021dori} use text encoders trained alongside their entire models. 
Despite these disadvantages due to the lack of text-related enhancements, our model still demonstrates state-of-the-art performance using the same video and text features (BERT~\Snowflake). This highlights the effectiveness of our design.

\subsection{Ablation Study}

Unless specified otherwise, we conducted the ablation study using our moment-DETR based model and evaluated on the test split of the IAW dataset.

\begin{table}[htb]
    \centering
    \scriptsize
    \newcolumntype{Z}{>{\centering\arraybackslash}X}
    \vspace{-0.2cm}
    \caption{Ablation results for different types of self-attention mask. The type name of the mask corresponds to that in~\cref{fig:mask}.}
    \begin{NiceTabularX}{\linewidth}{
            @{} l | *{3}{Z} | Z
        }[code-before = {\rowcolor{gray!25}{6}}]
        \toprule
        \Block{2-1}{Mask Type}
        & \Block{1-3}{R@1, IoU=}
        &
        &
        & \Block{2-1}{mIoU}
        \\
        \cmidrule(lr){2-4}
        & 0.3
        & 0.5
        & 0.7
        \\
        \midrule
        A
        & 36.25
        & 21.48
        & 8.655
        & 23.07
        \\
        B
        & 35.19
        & 20.14
        & 8.398
        & 22.37
        \\
        C
        & \underline{36.56}
        & \underline{21.77}
        & \underline{8.712}
        & \underline{23.43}
        \\
        D
        & \textbf{37.79}
        & \textbf{22.74}
        & \textbf{9.140}
        & \textbf{23.86}
        \\
        \bottomrule
    \end{NiceTabularX}
    \label{tab:mask_type}
    \vspace{-0.3cm}
\end{table}

\subsubsubsection{Choice of Self-Attention Masks.}
As show in~\cref{tab:mask_type}, the fact that type D actually outperforms type A suggests that those attentions crossing between different diagrams and different learnable queries tend to act as noise for the model. 
By masking them out, type D achieves the best performance. 
Notably, the implementation of type B, which essentially mirrors the single-query methods, has inferior performance than others, underscores the effeteness of constructing composite query via duplication.

\begin{table}[htb]
    \centering
    \scriptsize
    \newcolumntype{Z}{>{\centering\arraybackslash}X}
    \vspace{-0.2cm}
    \caption{Ablation results for different value choices for self-attention $V^\text{s}$ and cross-attention $V^\text{c}$. We use $\oplus$ to represent first concatenate then project back operation. $\dagger$: The self-attention output becomes $Q_c$, and $Q_p$ is then the learnable queries $\query$.}
    \begin{NiceTabularX}{\linewidth}{
            @{} l | Z | Z | *{3}{Z} | Z
        }[code-before = {\rowcolor{gray!25}{8}}]
        \toprule
        \Block{2-1}{\#}
        & \Block{2-1}{$V^{\text{s}}$}
        & \Block{2-1}{$V^{\text{c}}$}
        & \Block{1-3}{R@1, IoU=}
        &
        &
        & \Block{2-1}{mIoU}
        \\
        \cmidrule(lr){4-6}
        &
        &
        & 0.3
        & 0.5
        & 0.7
        \\
        \midrule
        1
        & $\diagramf$$^\dagger$
        & $\videof$$^\dagger$
        & 24.82
        & 12.43
        & 4.456
        & 16.34
        \\
        2
        & $\query$
        & $\videop$
        & 35.68
        & 20.85
        & 8.255
        & 23.29
        \\
        3
        & $\query$
        & $\videop+\videof$
        & \underline{36.65}
        & 20.99
        & 8.026
        & 23.18
        \\
        4
        & $\query+\diagramf$
        & $\videop+\videof$
        & 35.56
        & 20.51
        & \underline{8.769}
        & 22.81
        \\
        5
        & $\query$
        & $\videop\oplus\videof$
        & 36.33
        & \underline{21.37}
        & 8.283
        & \underline{23.37}
        \\
        6
        & $\query\oplus\diagramf$
        & $\videop\oplus\videof$
        & \textbf{37.79}
        & \textbf{22.74}
        & \textbf{9.140}
        & \textbf{23.86}
        \\
        \bottomrule
    \end{NiceTabularX}
    \label{tab:attention_value}
    \vspace{-0.3cm}
\end{table}

\subsubsubsection{Choice of Attention Values.}
By altering the input value feature for both self-attention $V^{\text{s}}$ and cross-attention $V^{\text{c}}$, we attempt to gain insight into the model's learning process for timespan predictions. As discussed in~\cref{sec:value}, setting $V^{\text{s}}$ to $\diagramf$ and $V^{\text{c}}$ to $\videof$, as shown in the first row of ~\cref{tab:attention_value}, suggests that the decoder operates within the feature space, producing an output that is a weighted combination of $\diagramf$ and $\videof$. Similarly, for the second line where $V^{\text{s}}=\diagramp$ and $V^{\text{c}}=\videop$, the decoder functions in the positional encoding space, leading to an output that integrates only $\diagramp$ and $\videop$. Because the timespan predictions are semantically more related to the positional encoding space, we hypothesize that the model can learn temporal information explicitly from those positional encoding. The fact that the first line outperforms random guessing in~\cref{tab:main_result} indicates the model's ability to also indirectly learn from the feature space. The comparison between the first two lines depicts the space where learning is more effective, with positional encoding space being the clear winner. The lines 2 and 4 show that simply adding the two inputs do not improve the performance. A more effective approach---as demonstrated in the sixth row---involves first concatenating then projecting, granting the model greater flexibility in merging these two spaces rather than relying on their sum, for information extraction. Lines 3 and 5 suggest that removing diagram features from the value feature of self-attention do not further improves the performance.

\begin{table}[htb]
    \centering
    \scriptsize
    \newcolumntype{Z}{>{\centering\arraybackslash}X}
    \vspace{-0.2cm}
    \caption{Ablation results for component designs.}
    \begin{NiceTabularX}{\linewidth}{
            @{} l | *{3}{Z} | Z
        }
        \toprule
        \Block{2-1}{Method}
        & \Block{1-3}{R@1, IoU=}
        &
        &
        & \Block{2-1}{mIoU}
        \\
        \cmidrule(lr){2-4}
        & 0.3
        & 0.5
        & 0.7
        \\
        \midrule
        Ours
        & \textbf{37.79}
        & \textbf{22.74}
        & \textbf{9.140}
        & \textbf{23.86}
        \\
        \quad w/o Diagram PE
        & 33.99
        & 19.42
        & 8.226
        & 21.63
        \\
        \quad w/o Contrastive Alignment
        & 33.82
        & 18.48
        & 7.112
        & 22.03
        \\
        \quad w/o Auxiliary Loss
        & 34.93
        & 19.11
        & 7.541
        & 22.00
        \\
        \quad w/o Average Video Length
        & 36.08
        & 21.19
        & 8.283
        & 22.92
        \\
        \bottomrule
    \end{NiceTabularX}
    \label{tab:component}
    \vspace{-0.3cm}
\end{table}

\subsubsubsection{Effect of Component Design.}
We also ablate on the design of other components in our model in~\cref{tab:component}. First, we found that incorporating order information into the inference model improves performance. This is achieved through diagram PE $\diagramp$, which ideally should highlight the diagonal in the similarity matrix $\diagramp \videop$, indicating the alignment between the two sequences. Second, contrastively aligning features prior to the decoder is vital. This necessity arises because video and image encoders we used are trained separately, leading to their feature representations laying in different spaces. Furthermore, our evaluation confirms the benefit of applying the same loss on the output of each layer (the auxiliary loss) in enhancing the overall performance. Last, due the limitation of 2D-TAN, it averages the length of video clip to 256 by interpolation. We find that doing this also helps improve performance, but our model can also handle varying duration using padding thanks to the nature of the attention mechanism.

\subsection{Analysis, Qualitative Results and Limitations}

\begin{figure}[htb]
  \centering
  \begin{subfigure}{0.49\linewidth}
    \includegraphics[width=\linewidth]{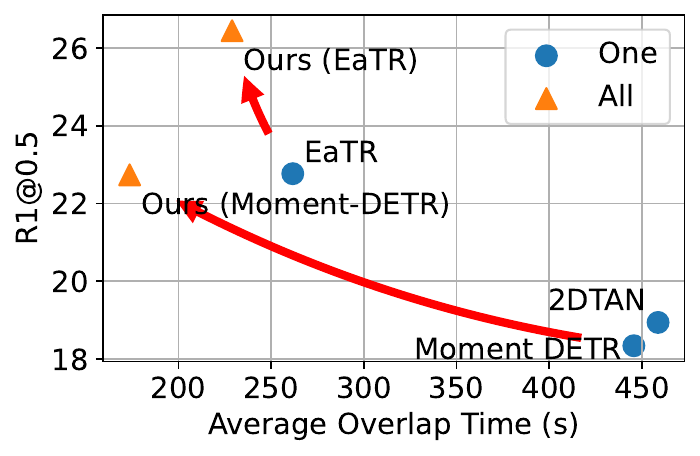}
    \label{fig:overlap}
  \end{subfigure}
  \hfill
  \begin{subfigure}{0.49\linewidth}
    \includegraphics[width=\linewidth]{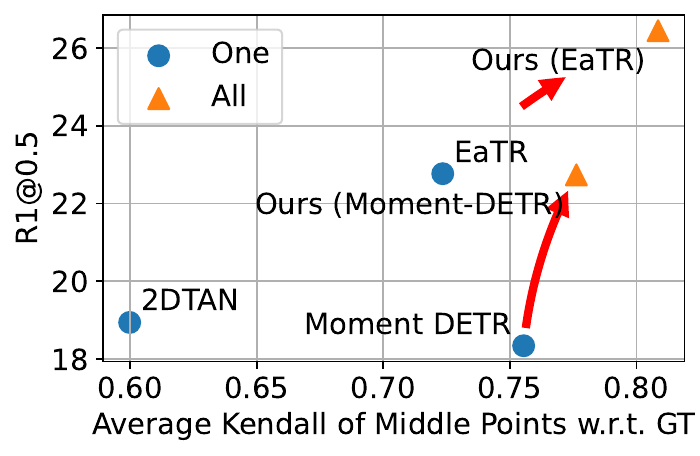}
    \label{fig:kendall}
  \end{subfigure}
  \vspace{-0.7cm}
  \caption{Two factors that affect the performance on IAW test.}
  \label{fig:analysis}
  \vspace{-0.64cm}
\end{figure}

\subsubsubsection{Timespan Analysis.} 
We investigated two factors that influence performance by analyzing the top-1 timespan predictions along with their ground truth, as shown in~\cref{fig:analysis}. 
The left diagram demonstrates the relationship between the average overlapping time among timespans for a sequence of queries and the performance (R1@0.5). 
The right diagram shows the Kendall's Tau coefficient between the middle points of predicted and ground truth timespans, where a higher score indicates a stronger positive correlation. 
It can be observed that superior models tend to have shorter overlapping times and higher correlation coefficients. 
By transitioning from single-query (One) to multi-query (All) using our approach, we consistently observe that as overlapping time decreases and the correlation coefficient increases, performance improves for both Moment-DETR and EaTR.

\begin{figure}[htb]
  \vspace{-0.2cm}
  \centering
  \begin{subfigure}{\linewidth}
    \includegraphics[width=\linewidth]{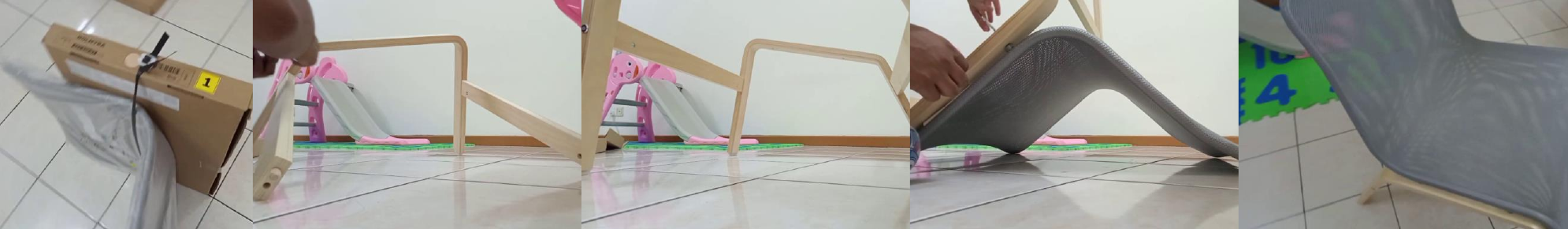}
    \caption{Assembly video from YouTube \href{https://www.youtube.com/watch?v=1czDviZ5vG0}{1czDviZ5vG0}.}
    \label{fig:ca-video}
  \end{subfigure}
  \\
  \begin{subfigure}{\linewidth}
    \includegraphics[width=\linewidth]{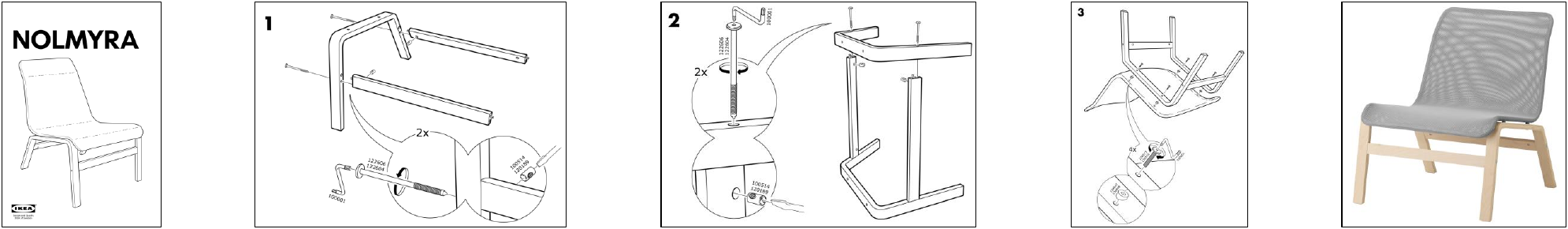}
    \caption{Step diagrams from Ikea furniture \href{https://www.ikea.com/au/en/p/nolmyra-easy-chair-birch-veneer-grey-10233607/}{10233607}.}
    \label{fig:ca-diagram}
  \end{subfigure}
  \\
  \begin{subfigure}{0.31\linewidth}
    \includegraphics[width=\linewidth]{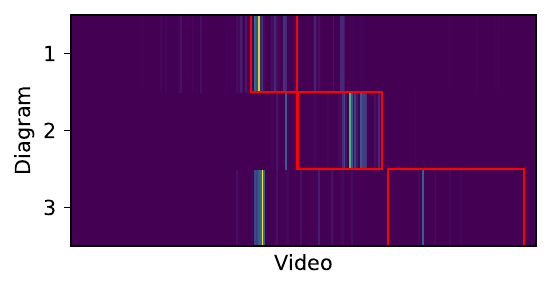}
    \caption{$\sigma(Q_c K_c \transpose)$}
    \label{fig:ca-f}
  \end{subfigure}
  \begin{subfigure}{0.335\linewidth}
    \includegraphics[width=\linewidth]{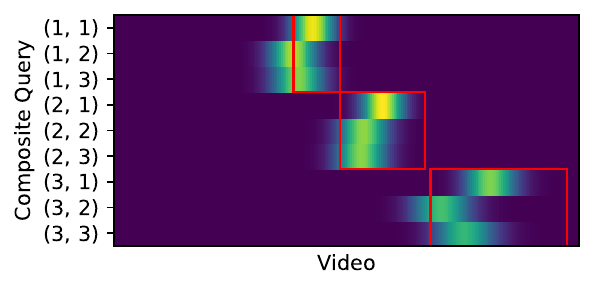}
    \caption{$\sigma(Q_p K_p \transpose)$}
    \label{fig:ca-p}
  \end{subfigure}
  \begin{subfigure}{0.335\linewidth}
    \includegraphics[width=\linewidth]{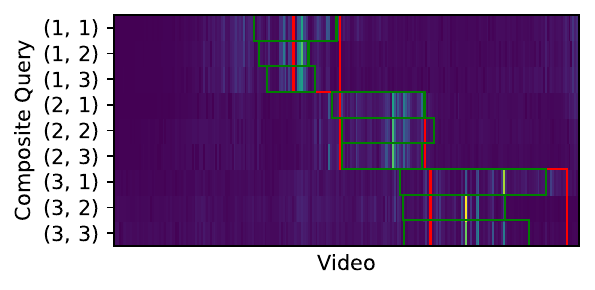}
    \caption{$\sigma(Q_c K_c \transpose + Q_p K_p \transpose)$}
    \label{fig:ca-add}
  \end{subfigure}
  \caption{Visualization of content and position jointly guided cross-attention for an example with three diagrams and three learnable queries ($M=3,K=3$). The top two rows (\subref{fig:ca-video},~\subref{fig:ca-diagram}) demonstrate the video and the corresponding step diagrams. The third row~(\subref{fig:ca-f},~\subref{fig:ca-p},~\subref{fig:ca-add}) visualizes each term and the outcome of~\cref{eq:a}, with ground truth timespans marked in red and predictions of each query highlighted in green. Composite queries are denoted as (diagram, learnable query). We use $\sigma$ to denote the softmax operation applied with scaling.}
  \label{fig:cross-attention}
  \vspace{-0.4cm}
\end{figure}

\subsubsubsection{Visualization of Cross Attention.}
To elucidate the operation of our joint content-and-position-guided cross-attention mechanism, we qualitatively analyze each term in~\cref{eq:a}. As~\cref{fig:ca-f} reveals, videos sourced from YouTube can frequently introduce noise, such as the distinct outlier slightly to the left of the center in the timeline depicted in the third row. Nonetheless, the positional guidance, illustrated in~\cref{fig:ca-p}, effectively self-corrects this misalignment, ensuring that the final predictions are not in the outlier location. Additionally, different queries tend to look at different locations. For example, the second and third queries focus on earlier positions compared to the first query. This tendency enables the model to navigate the temporal dimension effectively. Finally, the interplay between these dynamics culminates in jointly guided cross-attention, as illustrated in~\cref{fig:ca-add}, which significantly influences the final timespan predictions, showcasing the integrated effect of semantic and positional cues.

\begin{figure}[htb]
  \vspace{-0.2cm}
  \begin{subfigure}{\linewidth}
    \includegraphics[width=\linewidth]{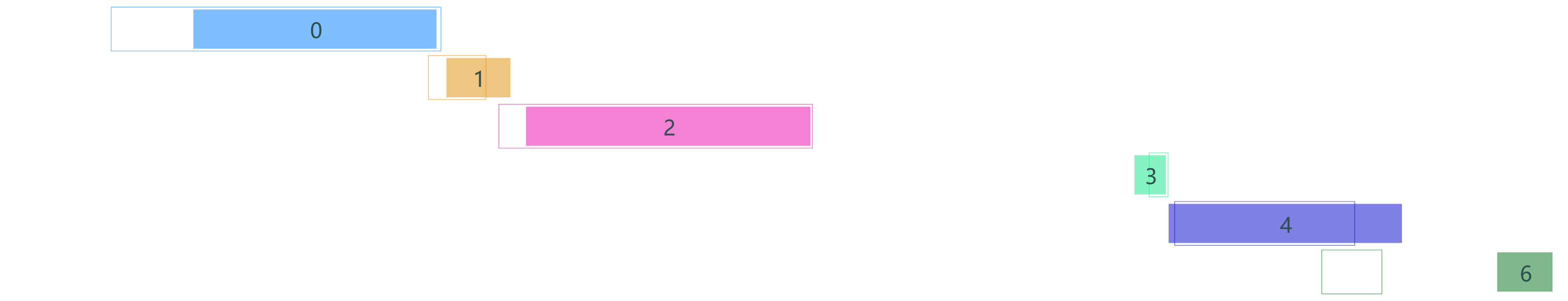}
    \caption{A successful grounding example between video \href{https://www.youtube.com/watch?v=VRobR2Yk1So}{VRobR2Yk1So} and furniture \href{https://www.ikea.com/au/en/p/mittback-trestle-birch-50459996/}{50459996}.}
    \label{fig:example-good}
  \end{subfigure}
  \\
  \begin{subfigure}{\linewidth}
    \includegraphics[width=\linewidth]{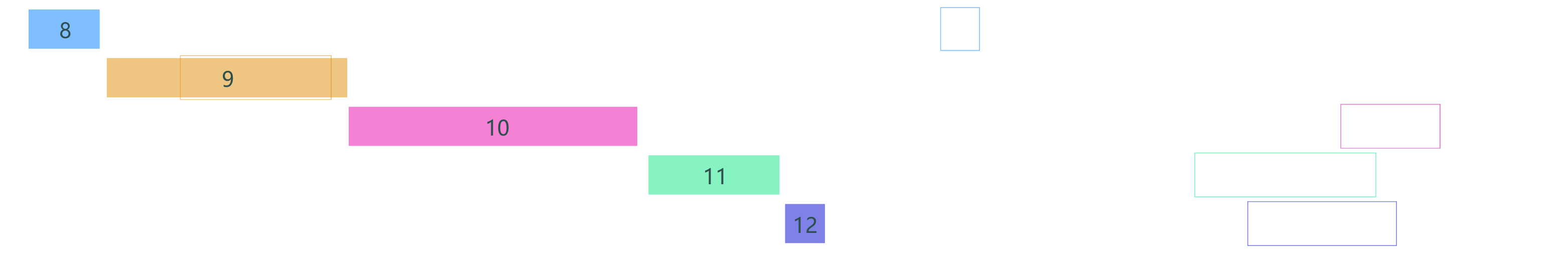}
    \caption{A failed grounding example between video \href{https://www.youtube.com/watch?v=ErgTQhbaIqg}{ErgTQhbaIqg} and furniture \href{https://www.ikea.com/au/en/p/eket-wall-mounted-shelving-unit-white-s99285810/}{s99285810}.}
    \label{fig:example-bad}
  \end{subfigure}
  \caption{Qualitative results. The horizontal axis represents the timeline of the video. Each row corresponds to a step diagram where the solid rectangle denotes the ground truth and top-1 timespan predictions are represented by bounding boxes with the same color. Notably, the video in~\subref{fig:example-bad} starts from step 8.}
  \label{fig:example}
  \vspace{-0.4cm}
\end{figure}

\subsubsubsection{Visualization of Examples and Limitations.}
As shown in~\cref{fig:example-good} our model accurately aligns the predicted timespans with the ground truth except for the last diagram. In contrast, \cref{fig:example-bad} illustrates a missed grounding result in which only one out of five predictions finds the correct location. This could be attributed to the diagram PE we introduce to the model. Notably, the video starts from step 8, leading to the majority of our predicted timespans occurring in the latter half of the video.

\section{Conclusion}
\label{sec:conclusion}

In this paper, we present a novel approach for the temporal grounding of instructional diagrams within unconstrained videos. Our method specifically addresses the challenges posed by the inherent sequential structure of instructional content, such as assembly manuals. We introduced composite queries by pairing diagram features with learnable positional queries, allowing our model to leverage both semantic and positional information effectively via content and position joint guided cross-attention. This design not only improves the accuracy of temporal grounding but also provides insights into the attention mechanisms for temporal grounding task.

There are several avenues for future research. For instance, further exploration into the handling of learnable queries and the integration of additional modalities (\eg, audio cues from videos) could enhance the model's robustness and accuracy. Additionally, extending our approach to other domains where sequential events are prevalent, such as medical procedure videos or educational content, presents an exciting opportunity for broader impact.

{\small
\bibliographystyle{ieee_fullname}
\bibliography{paper}

\begin{thebibliography}{10}\itemsep=-1pt

\bibitem{anne2017localizing}
Lisa Anne~Hendricks, Oliver Wang, Eli Shechtman, Josef Sivic, Trevor Darrell, and Bryan Russell.
\newblock Localizing moments in video with natural language.
\newblock In {\em Proceedings of the IEEE international conference on computer vision}, pages 5803--5812, 2017.

\bibitem{bao2021dense}
Peijun Bao, Qian Zheng, and Yadong Mu.
\newblock Dense events grounding in video.
\newblock In {\em Proceedings of the AAAI Conference on Artificial Intelligence}, volume~35, pages 920--928, 2021.

\bibitem{ben2021ikea}
Yizhak Ben-Shabat, Xin Yu, Fatemeh Saleh, Dylan Campbell, Cristian Rodriguez-Opazo, Hongdong Li, and Stephen Gould.
\newblock The {Ikea} {ASM} dataset: Understanding people assembling furniture through actions, objects and pose.
\newblock In {\em Proceedings of the IEEE/CVF Winter Conference on Applications of Computer Vision}, pages 847--859, 2021.

\bibitem{carion2020end}
Nicolas Carion, Francisco Massa, Gabriel Synnaeve, Nicolas Usunier, Alexander Kirillov, and Sergey Zagoruyko.
\newblock End-to-end object detection with transformers.
\newblock In {\em European conference on computer vision}, pages 213--229. Springer, 2020.

\bibitem{carreira2017quo}
Joao Carreira and Andrew Zisserman.
\newblock Quo vadis, action recognition? a new model and the kinetics dataset.
\newblock In {\em proceedings of the IEEE Conference on Computer Vision and Pattern Recognition}, pages 6299--6308, 2017.

\bibitem{chen2018temporally}
Jingyuan Chen, Xinpeng Chen, Lin Ma, Zequn Jie, and Tat-Seng Chua.
\newblock Temporally grounding natural sentence in video.
\newblock In {\em Proceedings of the 2018 conference on empirical methods in natural language processing}, pages 162--171, 2018.

\bibitem{damen2022rescaling}
Dima Damen, Hazel Doughty, Giovanni~Maria Farinella, Antonino Furnari, Evangelos Kazakos, Jian Ma, Davide Moltisanti, Jonathan Munro, Toby Perrett, Will Price, et~al.
\newblock Rescaling egocentric vision: Collection, pipeline and challenges for epic-kitchens-100.
\newblock {\em International Journal of Computer Vision}, pages 1--23, 2022.

\bibitem{devlin2018bert}
Jacob Devlin, Ming-Wei Chang, Kenton Lee, and Kristina Toutanova.
\newblock Bert: Pre-training of deep bidirectional transformers for language understanding.
\newblock {\em arXiv preprint arXiv:1810.04805}, 2018.

\bibitem{gao2017tall}
Jiyang Gao, Chen Sun, Zhenheng Yang, and Ram Nevatia.
\newblock Tall: Temporal activity localization via language query.
\newblock In {\em Proceedings of the IEEE international conference on computer vision}, pages 5267--5275, 2017.

\bibitem{ghosh2019excl}
Soham Ghosh, Anuva Agarwal, Zarana Parekh, and Alexander Hauptmann.
\newblock {E}x{CL}: {E}xtractive {C}lip {L}ocalization {U}sing {N}atural {L}anguage {D}escriptions.
\newblock In Jill Burstein, Christy Doran, and Thamar Solorio, editors, {\em Proceedings of the 2019 Conference of the North {A}merican Chapter of the Association for Computational Linguistics: Human Language Technologies, Volume 1 (Long and Short Papers)}, pages 1984--1990, Minneapolis, Minnesota, June 2019. Association for Computational Linguistics.

\bibitem{jang2023knowing}
Jinhyun Jang, Jungin Park, Jin Kim, Hyeongjun Kwon, and Kwanghoon Sohn.
\newblock Knowing where to focus: Event-aware transformer for video grounding.
\newblock In {\em Proceedings of the IEEE/CVF International Conference on Computer Vision}, pages 13846--13856, 2023.

\bibitem{jiang2022semi}
Xun Jiang, Xing Xu, Jingran Zhang, Fumin Shen, Zuo Cao, and Heng~Tao Shen.
\newblock Semi-supervised video paragraph grounding with contrastive encoder.
\newblock In {\em Proceedings of the IEEE/CVF Conference on Computer Vision and Pattern Recognition}, pages 2466--2475, 2022.

\bibitem{wang2023ms}
Wang Jing, Aixin Sun, Hao Zhang, and Xiaoli Li.
\newblock {MS}-{DETR}: Natural language video localization with sampling moment-moment interaction.
\newblock In Anna Rogers, Jordan Boyd-Graber, and Naoaki Okazaki, editors, {\em Proceedings of the 61st Annual Meeting of the Association for Computational Linguistics (Volume 1: Long Papers)}, pages 1387--1400, Toronto, Canada, July 2023. Association for Computational Linguistics.

\bibitem{shimomoto2022towards}
Erica Kido~Shimomoto, Edison Marrese-Taylor, Hiroya Takamura, Ichiro Kobayashi, Hideki Nakayama, and Yusuke Miyao.
\newblock Towards parameter-efficient integration of pre-trained language models in temporal video grounding.
\newblock In Anna Rogers, Jordan Boyd-Graber, and Naoaki Okazaki, editors, {\em Findings of the Association for Computational Linguistics: ACL 2023}, pages 13101--13123, Toronto, Canada, July 2023. Association for Computational Linguistics.

\bibitem{kim2023self}
Jihwan Kim, Miso Lee, and Jae-Pil Heo.
\newblock Self-feedback {DETR} for temporal action detection.
\newblock In {\em Proceedings of the IEEE/CVF International Conference on Computer Vision}, pages 10286--10296, 2023.

\bibitem{Krishna_2017_ICCV}
Ranjay Krishna, Kenji Hata, Frederic Ren, Li Fei-Fei, and Juan Carlos~Niebles.
\newblock Dense-captioning events in videos.
\newblock In {\em Proceedings of the IEEE International Conference on Computer Vision (ICCV)}, Oct 2017.

\bibitem{kuhn1955hungarian}
Harold~W Kuhn.
\newblock The hungarian method for the assignment problem.
\newblock {\em Naval research logistics quarterly}, 2(1-2):83--97, 1955.

\bibitem{lei2021detecting}
Jie Lei, Tamara~L Berg, and Mohit Bansal.
\newblock Detecting moments and highlights in videos via natural language queries.
\newblock {\em Advances in Neural Information Processing Systems}, 34:11846--11858, 2021.

\bibitem{lin2023univtg}
Kevin~Qinghong Lin, Pengchuan Zhang, Joya Chen, Shraman Pramanick, Difei Gao, Alex~Jinpeng Wang, Rui Yan, and Mike~Zheng Shou.
\newblock Univtg: Towards unified video-language temporal grounding.
\newblock In {\em Proceedings of the IEEE/CVF International Conference on Computer Vision}, pages 2794--2804, 2023.

\bibitem{liu2022end}
Xiaolong Liu, Qimeng Wang, Yao Hu, Xu Tang, Shiwei Zhang, Song Bai, and Xiang Bai.
\newblock End-to-end temporal action detection with transformer.
\newblock {\em IEEE Transactions on Image Processing}, 31:5427--5441, 2022.

\bibitem{loshchilov2017decoupled}
Ilya Loshchilov and Frank Hutter.
\newblock Decoupled weight decay regularization.
\newblock {\em arXiv preprint arXiv:1711.05101}, 2017.

\bibitem{meng2021conditional}
Depu Meng, Xiaokang Chen, Zejia Fan, Gang Zeng, Houqiang Li, Yuhui Yuan, Lei Sun, and Jingdong Wang.
\newblock Conditional {DETR} for fast training convergence.
\newblock In {\em Proceedings of the IEEE/CVF International Conference on Computer Vision}, pages 3651--3660, 2021.

\bibitem{miech2019howto100m}
Antoine Miech, Dimitri Zhukov, Jean-Baptiste Alayrac, Makarand Tapaswi, Ivan Laptev, and Josef Sivic.
\newblock Howto100m: Learning a text-video embedding by watching hundred million narrated video clips.
\newblock In {\em Proceedings of the IEEE/CVF international conference on computer vision}, pages 2630--2640, 2019.

\bibitem{moon2023correlation}
WonJun Moon, Sangeek Hyun, SuBeen Lee, and Jae-Pil Heo.
\newblock Correlation-guided query-dependency calibration in video representation learning for temporal grounding.
\newblock {\em arXiv preprint arXiv:2311.08835}, 2023.

\bibitem{moon2023query}
WonJun Moon, Sangeek Hyun, SangUk Park, Dongchan Park, and Jae-Pil Heo.
\newblock Query-dependent video representation for moment retrieval and highlight detection.
\newblock In {\em Proceedings of the IEEE/CVF Conference on Computer Vision and Pattern Recognition}, pages 23023--23033, 2023.

\bibitem{oquab2023dinov2}
Maxime Oquab, Timoth{\'e}e Darcet, Th{\'e}o Moutakanni, Huy Vo, Marc Szafraniec, Vasil Khalidov, Pierre Fernandez, Daniel Haziza, Francisco Massa, Alaaeldin El-Nouby, et~al.
\newblock Dinov2: Learning robust visual features without supervision.
\newblock {\em arXiv preprint arXiv:2304.07193}, 2023.

\bibitem{regneri2013grounding}
Michaela Regneri, Marcus Rohrbach, Dominikus Wetzel, Stefan Thater, Bernt Schiele, and Manfred Pinkal.
\newblock Grounding action descriptions in videos.
\newblock {\em Transactions of the Association for Computational Linguistics}, 1:25--36, 2013.

\bibitem{rezatofighi2019generalized}
Hamid Rezatofighi, Nathan Tsoi, JunYoung Gwak, Amir Sadeghian, Ian Reid, and Silvio Savarese.
\newblock Generalized intersection over union: A metric and a loss for bounding box regression.
\newblock In {\em Proceedings of the IEEE/CVF conference on computer vision and pattern recognition}, pages 658--666, 2019.

\bibitem{rodriguez2020proposal}
Cristian Rodriguez, Edison Marrese-Taylor, Fatemeh~Sadat Saleh, Hongdong Li, and Stephen Gould.
\newblock Proposal-free temporal moment localization of a natural-language query in video using guided attention.
\newblock In {\em Proceedings of the IEEE/CVF winter conference on applications of computer vision}, pages 2464--2473, 2020.

\bibitem{rodriguez2021dori}
Cristian Rodriguez-Opazo, Edison Marrese-Taylor, Basura Fernando, Hongdong Li, and Stephen Gould.
\newblock Dori: Discovering object relationships for moment localization of a natural language query in a video.
\newblock In {\em Proceedings of the IEEE/CVF Winter Conference on Applications of Computer Vision}, pages 1079--1088, 2021.

\bibitem{rodriguez2021locformer}
Cristian Rodriguez-Opazo, Edison Marrese-Taylor, Basura Fernando, Hiroya Takamura, and Qi Wu.
\newblock Memory-efficient temporal moment localization in long videos.
\newblock In Andreas Vlachos and Isabelle Augenstein, editors, {\em Proceedings of the 17th Conference of the European Chapter of the Association for Computational Linguistics}, pages 1909--1924, Dubrovnik, Croatia, May 2023. Association for Computational Linguistics.

\bibitem{rodriguez-opazo-etal-2023-memory}
Cristian Rodriguez-Opazo, Edison Marrese-Taylor, Basura Fernando, Hiroya Takamura, and Qi Wu.
\newblock Memory-efficient temporal moment localization in long videos.
\newblock In {\em Proceedings of the 17th Conference of the European Chapter of the Association for Computational Linguistics}. Association for Computational Linguistics, May 2023.

\bibitem{rohrbach2012script}
Marcus Rohrbach, Michaela Regneri, Mykhaylo Andriluka, Sikandar Amin, Manfred Pinkal, and Bernt Schiele.
\newblock Script data for attribute-based recognition of composite activities.
\newblock In {\em Computer Vision--ECCV 2012: 12th European Conference on Computer Vision, Florence, Italy, October 7-13, 2012, Proceedings, Part I 12}, pages 144--157. Springer, 2012.

\bibitem{shi2021end}
Fengyuan Shi, Limin Wang, and Weilin Huang.
\newblock End-to-end dense video grounding via parallel regression.
\newblock {\em arXiv preprint arXiv:2109.11265}, 2021.

\bibitem{sun2024diversifying}
Xiaolong Sun, Liushuai Shi, Le Wang, Sanping Zhou, Kun Xia, Yabing Wang, and Gang Hua.
\newblock Diversifying query: Region-guided transformer for temporal sentence grounding.
\newblock {\em arXiv preprint arXiv:2406.00143}, 2024.

\bibitem{tan2023hierarchical}
Chaolei Tan, Zihang Lin, Jian-Fang Hu, Wei-Shi Zheng, and Jianhuang Lai.
\newblock Hierarchical semantic correspondence networks for video paragraph grounding.
\newblock In {\em Proceedings of the IEEE/CVF Conference on Computer Vision and Pattern Recognition}, pages 18973--18982, 2023.

\bibitem{tang2019coin}
Yansong Tang, Dajun Ding, Yongming Rao, Yu Zheng, Danyang Zhang, Lili Zhao, Jiwen Lu, and Jie Zhou.
\newblock Coin: A large-scale dataset for comprehensive instructional video analysis.
\newblock In {\em Proceedings of the IEEE/CVF Conference on Computer Vision and Pattern Recognition}, pages 1207--1216, 2019.

\bibitem{vaswani2017attention}
Ashish Vaswani, Noam Shazeer, Niki Parmar, Jakob Uszkoreit, Llion Jones, Aidan~N Gomez, {\L}ukasz Kaiser, and Illia Polosukhin.
\newblock Attention is all you need.
\newblock {\em Advances in neural information processing systems}, 30, 2017.

\bibitem{wang2023videomae}
Limin Wang, Bingkun Huang, Zhiyu Zhao, Zhan Tong, Yinan He, Yi Wang, Yali Wang, and Yu Qiao.
\newblock Videomae v2: Scaling video masked autoencoders with dual masking.
\newblock In {\em Proceedings of the IEEE/CVF Conference on Computer Vision and Pattern Recognition}, pages 14549--14560, 2023.

\bibitem{wang2019youmakeup}
Weiying Wang, Yongcheng Wang, Shizhe Chen, and Qin Jin.
\newblock Youmakeup: A large-scale domain-specific multimodal dataset for fine-grained semantic comprehension.
\newblock In {\em Proceedings of the 2019 Conference on Empirical Methods in Natural Language Processing and the 9th International Joint Conference on Natural Language Processing (EMNLP-IJCNLP)}, pages 5133--5143, 2019.

\bibitem{woo2022explore}
Sangmin Woo, Jinyoung Park, Inyong Koo, Sumin Lee, Minki Jeong, and Changick Kim.
\newblock Explore-and-match: Bridging proposal-based and proposal-free with transformer for sentence grounding in videos.
\newblock {\em arXiv preprint arXiv:2201.10168}, 2022.

\bibitem{xiao2024bridging}
Yicheng Xiao, Zhuoyan Luo, Yong Liu, Yue Ma, Hengwei Bian, Yatai Ji, Yujiu Yang, and Xiu Li.
\newblock Bridging the gap: A unified video comprehension framework for moment retrieval and highlight detection.
\newblock In {\em Proceedings of the IEEE/CVF Conference on Computer Vision and Pattern Recognition}, pages 18709--18719, 2024.

\bibitem{yang2024task}
Jin Yang, Ping Wei, Huan Li, and Ziyang Ren.
\newblock Task-driven exploration: Decoupling and inter-task feedback for joint moment retrieval and highlight detection.
\newblock In {\em Proceedings of the IEEE/CVF Conference on Computer Vision and Pattern Recognition}, pages 18308--18318, 2024.

\bibitem{yuan2019semantic}
Yitian Yuan, Lin Ma, Jingwen Wang, Wei Liu, and Wenwu Zhu.
\newblock Semantic conditioned dynamic modulation for temporal sentence grounding in videos.
\newblock {\em Advances in Neural Information Processing Systems}, 32, 2019.

\bibitem{yuan2019find}
Yitian Yuan, Tao Mei, and Wenwu Zhu.
\newblock To find where you talk: Temporal sentence localization in video with attention based location regression.
\newblock In {\em Proceedings of the AAAI Conference on Artificial Intelligence}, volume~33, pages 9159--9166, 2019.

\bibitem{zhang2019man}
Da Zhang, Xiyang Dai, Xin Wang, Yuan-Fang Wang, and Larry~S Davis.
\newblock Man: Moment alignment network for natural language moment retrieval via iterative graph adjustment.
\newblock In {\em Proceedings of the IEEE/CVF Conference on Computer Vision and Pattern Recognition}, pages 1247--1257, 2019.

\bibitem{zhang2022efficient}
Frederic~Z Zhang, Dylan Campbell, and Stephen Gould.
\newblock Efficient two-stage detection of human-object interactions with a novel unary-pairwise transformer.
\newblock In {\em Proceedings of the IEEE/CVF Conference on Computer Vision and Pattern Recognition}, pages 20104--20112, 2022.

\bibitem{zhang2023exploring}
Frederic~Z Zhang, Yuhui Yuan, Dylan Campbell, Zhuoyao Zhong, and Stephen Gould.
\newblock Exploring predicate visual context in detecting of human-object interactions.
\newblock In {\em Proceedings of the IEEE/CVF International Conference on Computer Vision}, pages 10411--10421, 2023.

\bibitem{zhang2021parallel}
Hao Zhang, Aixin Sun, Wei Jing, Liangli Zhen, Joey~Tianyi Zhou, and Siow Mong~Rick Goh.
\newblock Parallel attention network with sequence matching for video grounding.
\newblock In Chengqing Zong, Fei Xia, Wenjie Li, and Roberto Navigli, editors, {\em Findings of the Association for Computational Linguistics: ACL-IJCNLP 2021}, pages 776--790, Online, Aug. 2021. Association for Computational Linguistics.

\bibitem{zhang2020span}
Hao Zhang, Aixin Sun, Wei Jing, and Joey~Tianyi Zhou.
\newblock Span-based localizing network for natural language video localization.
\newblock In Dan Jurafsky, Joyce Chai, Natalie Schluter, and Joel Tetreault, editors, {\em Proceedings of the 58th Annual Meeting of the Association for Computational Linguistics}, pages 6543--6554, Online, July 2020. Association for Computational Linguistics.

\bibitem{zhang2023temporal}
Hao Zhang, Aixin Sun, Wei Jing, and Joey~Tianyi Zhou.
\newblock Temporal sentence grounding in videos: A survey and future directions.
\newblock {\em IEEE Transactions on Pattern Analysis and Machine Intelligence}, 2023.

\bibitem{zhang2023aligning}
Jiahao Zhang, Anoop Cherian, Yanbin Liu, Yizhak Ben-Shabat, Cristian Rodriguez, and Stephen Gould.
\newblock Aligning step-by-step instructional diagrams to video demonstrations.
\newblock In {\em Proceedings of the IEEE/CVF Conference on Computer Vision and Pattern Recognition}, pages 2483--2492, 2023.

\bibitem{zhang2020learning}
Songyang Zhang, Houwen Peng, Jianlong Fu, and Jiebo Luo.
\newblock Learning 2d temporal adjacent networks for moment localization with natural language.
\newblock In {\em Proceedings of the AAAI Conference on Artificial Intelligence}, volume~34, pages 12870--12877, 2020.

\bibitem{zhou2018towards}
Luowei Zhou, Chenliang Xu, and Jason Corso.
\newblock Towards automatic learning of procedures from web instructional videos.
\newblock In {\em Proceedings of the AAAI Conference on Artificial Intelligence}, volume~32, 2018.

\end{thebibliography}
}

\end{document}


\title{Temporally Grounding Instructional Diagrams in Unconstrained Videos Supplementary Material}

\author{Jiahao Zhang$^{1}$\quad
Frederic Z. Zhang$^{2}$\quad
Cristian Rodriguez$^{2}$\\
Yizhak Ben-Shabat$^{1,3}$\quad
Anoop Cherian$^{4}$\quad
Stephen Gould$^{1}$\\
$^1$The Australian National University\quad
$^2$The Australian Institute for Machine Learning\\
$^3$Technion Israel Institute of Technology\quad
$^4$Mitsubishi Electric Research Labs\\
{\tt\small $^1$\{first.last\}@anu.edu.au}\quad
{\tt\small $^2$\{first.last\}@adelaide.edu.au}\\
{\tt\small $^3$sitzikbs@gmail.com}\quad
{\tt\small $^4$cherian@merl.com}
}
\maketitle

\appendix

\section{Model Details}

\subsection{Contrastive Alignment}
Following Zhang \etal~\cite{zhang2023aligning}, contrastive alignment is applied to project video features $\videof$ and diagram features $\diagramf$ into a unified $D$-dimensional space such that the corresponding features are close to each other and irrelevant features are pushed far away from each other. Specifically, first encode the raw data into intermediate features using dedicated encoders. Second, concatenate them with the sinusoidal progress rate feature (SPRF)~\cite{zhang2023aligning}. Last, project them into same space and apply the video-manual contrastive loss (loss B~\cite{zhang2023aligning}) for pretraining. Notably, the background video features (which do not have corresponding diagrams) are ignored at this stage.

\subsection{Normalized Positional Encoding}
Due to the fact that the self-attention mechanism in Transformers~\cite{vaswani2017attention} is permutation invariant, it is essential for these models, including ours, to incorporate position information into each token. However, given the significant variation in the length of video clips $N$ (ranging from 25 to 2976) and the number of step diagrams $M$ (ranging from 2 to 79) found in manuals across our dataset, we need to adjust the positional encoding to make it independent of the sequence length. Formally, we have
\begin{gather}
\begin{aligned}
    PE^\text{V}_{(i,2t)}&=\sin\left(\frac{2\pi i/N}{10,000^{2t/d}}\right), \\
    PE^\text{V}_{(i,2t+1)}&=\cos\left(\frac{2\pi i/N}{10,000^{2t/d}}\right), \\
    PE^\text{I}_{(j,2t)}&=\sin\left(\frac{2\pi j/M}{10,000^{2t/d}}\right), \\
    PE^\text{I}_{(j,2t+1)}&=\cos\left(\frac{2\pi j/M}{10,000^{2t/d}}\right),
\end{aligned}
\end{gather}
where $i$ represents the index of the video clip in a video. Similarly, $j$ is the index of a step diagram in a manual book. Conceptually, we normalize the position \wrt their total length and then re-scale them to $[0, 2\pi]$. For convenience, the positional encodings for the video and diagram are denoted by $\videop\in\mathbb{R}^{N\times D}$ and $\diagramp\in\mathbb{R}^{M\times D}$, respectively.

\begin{table*}[tb]
    \centering
    \scriptsize
    \newcolumntype{Z}{>{\centering\arraybackslash}X}
    \caption{Details of the image encoder used for extracting diagram features in IAW.}
    \begin{NiceTabularX}{\linewidth}{
            @{} l | Z *{5}{|c}
        }
        \toprule
        Encoder
        & Arch.
        & Pretrained
        & Dim.
        & Norm mean
        & Norm std
        & Image size
        \\
        \midrule
        DINOv2~\cite{oquab2023dinov2}
        & ViT-G/14
        & LVD-142M~\cite{oquab2023dinov2}
        & 1536
        & 0.485, 0.456, 0.406
        & 0.229, 0.224, 0.225
        & 224*224
        \\
        \bottomrule
    \end{NiceTabularX}
    \label{tab:image-encoder}
\end{table*}

\begin{table*}[tb]
    \centering
    \scriptsize
    \newcolumntype{Z}{>{\centering\arraybackslash}X}
    \caption{Details of the video encoder used for extracting video features in IAW.}
    \begin{NiceTabularX}{\linewidth}{
            @{} l | Z | c *{4}{|Z}
        }
        \toprule
        Encoder
        & Arch.
        & Pretrained
        & Dim.
        & FPS
        & \#Frames
        & Frame size
        \\
        \midrule
        VideoMAEv2\cite{wang2023videomae}
        & ViT-G/14
        & UnlabeledHybrid\cite{wang2023videomae}
        & 1408
        & 10
        & 16
        & 224*224
        \\
        \bottomrule
    \end{NiceTabularX}
    \label{tab:video-encoder}
\end{table*}

\begin{table}[tb]
    \centering
    \scriptsize
    \newcolumntype{Z}{>{\centering\arraybackslash}X}
    \caption{Hyperparameters used in experiments for each datasets.}
    \begin{NiceTabularX}{\linewidth}{
            @{} l | Z | Z | Z
        }
        \toprule
        Name
        & IAW
        & YouCook2
        & ActivityNet Caption
        \\
        \midrule
        Learning rate
        & $10^{-4}$
        & $5\times 10^{-5}$
        & $10^{-5}$
        \\
        Weight decay
        & \Block{1-3}{$10^{-4}$}
        \\
        Betas
        & \Block{1-3}{$[0.9, 0.95]$}
        \\
        Scheduler
        & \Block{1-3}{MultiStepLR}
        \\
        Scheduler milestones
        & \Block{1-3}{$[3, 7, 15, 30, 50]$}
        \\
        Scheduler gamma
        & \Block{1-3}{0.5}
        \\
        \midrule
        Batch size
        & \Block{1-3}{16}
        \\
        Sampler window size
        & \Block{1-3}{$[128, 256, 512, 1024, \infty]$}
        \\
        Sampler stride
        & \Block{1-3}{$[32, 64, 128, 256, \infty]$}
        \\
        Video feature average length
        & \Block{1-3}{256}
        \\
        \midrule
        \# Queries
        & \Block{1-3}{3}
        \\
        Model dim
        & \Block{1-3}{768}
        \\
        Dropout
        & \Block{1-3}{0.1}
        \\
        Cost weight span L1 
        & \Block{1-3}{1}
        \\
        Cost weight span GIoU
        & \Block{1-3}{1}
        \\
        Cost weight score
        & \Block{1-3}{1}
        \\
        Loss weight span L1 
        & \Block{1-3}{1}
        \\
        Loss weight span GIoU
        & \Block{1-3}{1}
        \\
        Loss weight score
        & \Block{1-3}{1}
        \\
        Temperature init
        & \Block{1-3}{0.07}
        \\
        \bottomrule
    \end{NiceTabularX}
    \label{tab:hp}
\end{table}

\section{Experiment Details}

\subsection{Setup}
\label{sec:setup}

The essential configurations for the image encoder and the video encoder are detailed in~\cref{tab:image-encoder,tab:video-encoder}, respectively. Additionally, the key hyperparameters are presented in~\cref{tab:hp}.

\subsection{Modifications on Other Models}

In the TSGV models selected for comparison, the text encoder is removed; instead, diagram features are utilized for modality fusion with video features, mirroring those employed in our model. 
Specifically, for models such as 2D-TAN~\cite{zhang2020learning} and LVTR~\cite{woo2022explore}, the sentence features are directly substituted with the diagram features. 
Meanwhile, in the case of Moment-DETR~\cite{lei2021detecting} and EaTR~\cite{jang2023knowing}, rather than incorporating word-level features, a single diagram feature is concatenated with video features and input into the encoder. 
EaTR also changed the decoder where they introduced Gated Fusion Transformer Layer to replace the first layer.
The output from that layer is $\mathbf{C}\in\reals^{K\times d}$, we then treat $\mathbf{C}$ as the learnable queries in our paper.
Exhaustively pair up $\mathbf{C}$ and the diagram features for the rest $T-1$ Transformer Decoder layers, we enabled EaTR to ground multiple queries simultaneously. 
In particular, there are 10 learnable queries for Moment-DETR and EaTR, and 80 for LVTR.

\subsection{Dataset Bias}

\begin{figure}[tb]
  \centering
  \begin{subfigure}{0.325\linewidth}
    \includegraphics[width=\linewidth]{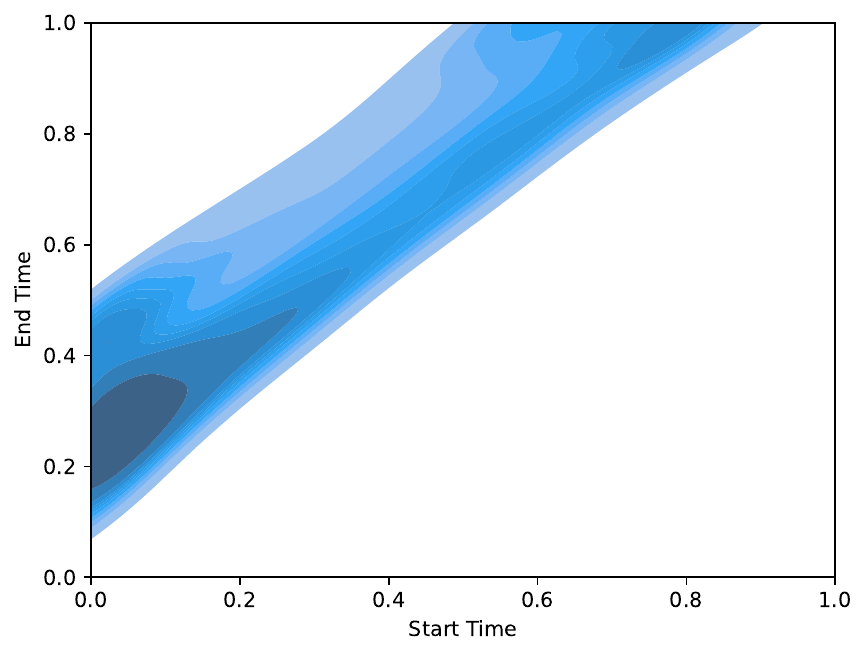}
    \caption{Charades STA Test}
    \label{fig:charades_val}
  \end{subfigure}
  \hfill
  \begin{subfigure}{0.325\linewidth}
    \includegraphics[width=\linewidth]{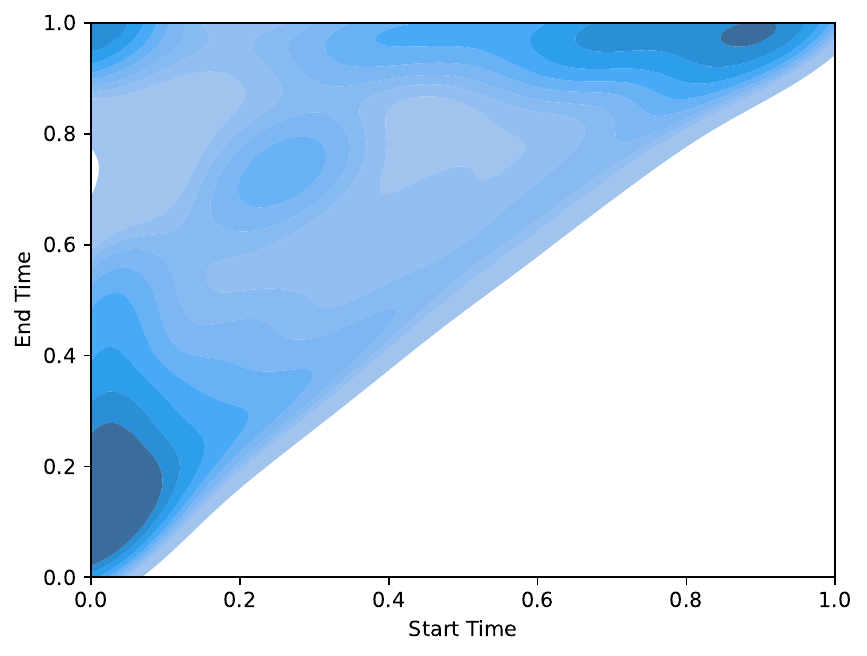}
    \caption{ANet-Caption Val}
    \label{fig:anet_val}
  \end{subfigure}
  \hfill
  \begin{subfigure}{0.325\linewidth}
    \includegraphics[width=\linewidth]{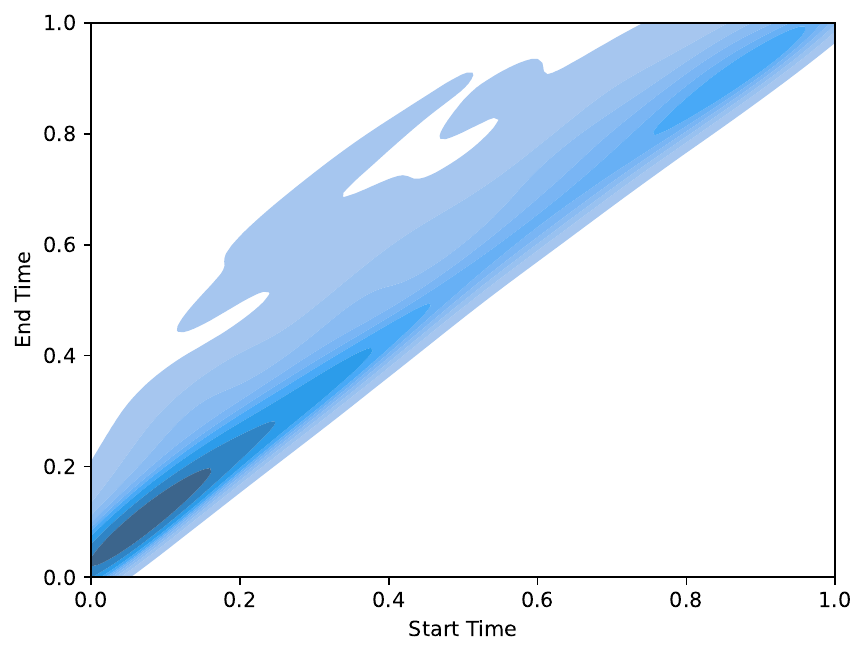}
    \caption{TACoS Val}
    \label{fig:tacos_val}
  \end{subfigure}
  \\
  \begin{subfigure}{0.325\linewidth}
    \includegraphics[width=\linewidth]{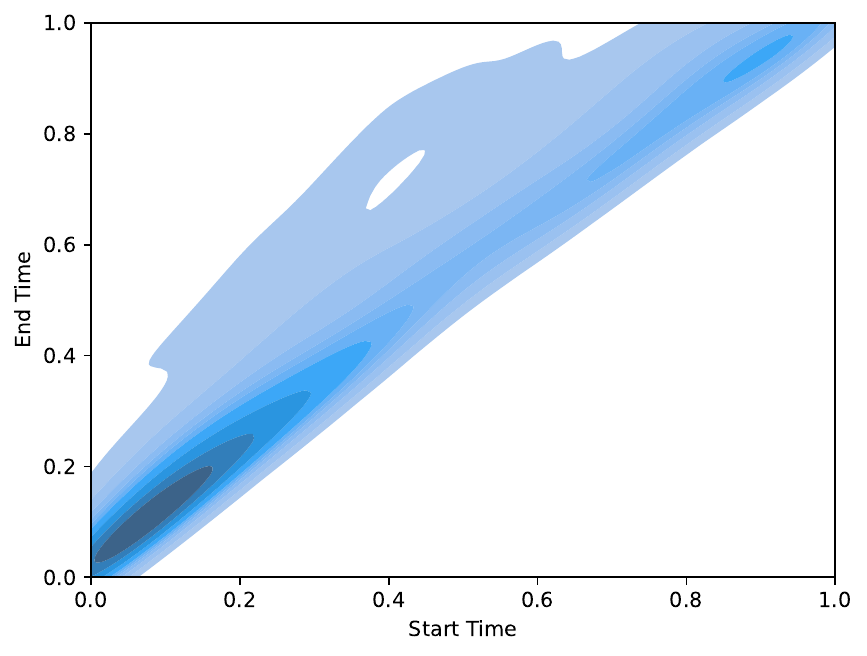}
    \caption{TACoS Test}
    \label{fig:tacos_test}
  \end{subfigure}
  \hfill
  \begin{subfigure}{0.325\linewidth}
    \includegraphics[width=\linewidth]{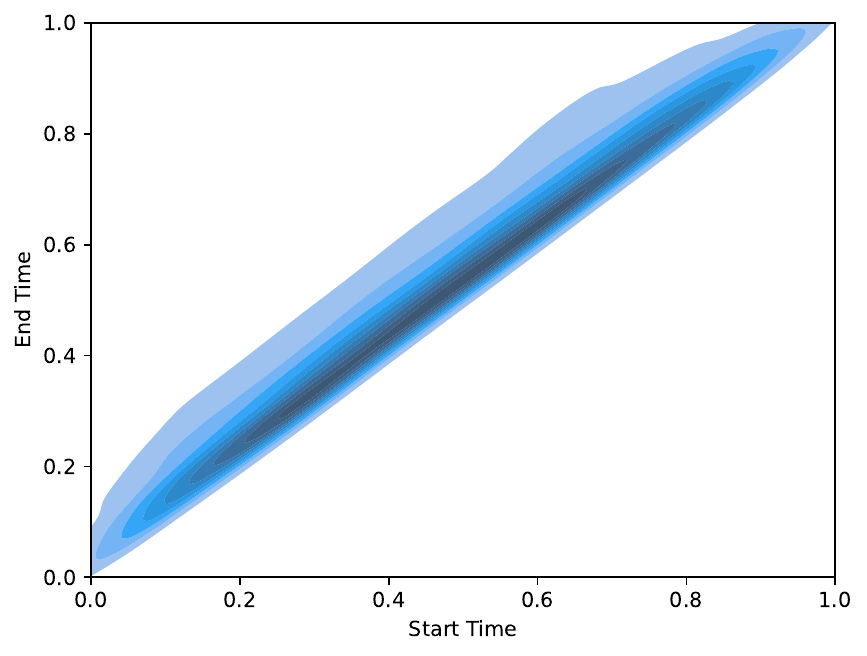}
    \caption{YouCookII Val}
    \label{fig:youcook2_val}
  \end{subfigure}
  \hfill
  \begin{subfigure}{0.325\linewidth}
    \includegraphics[width=\linewidth]{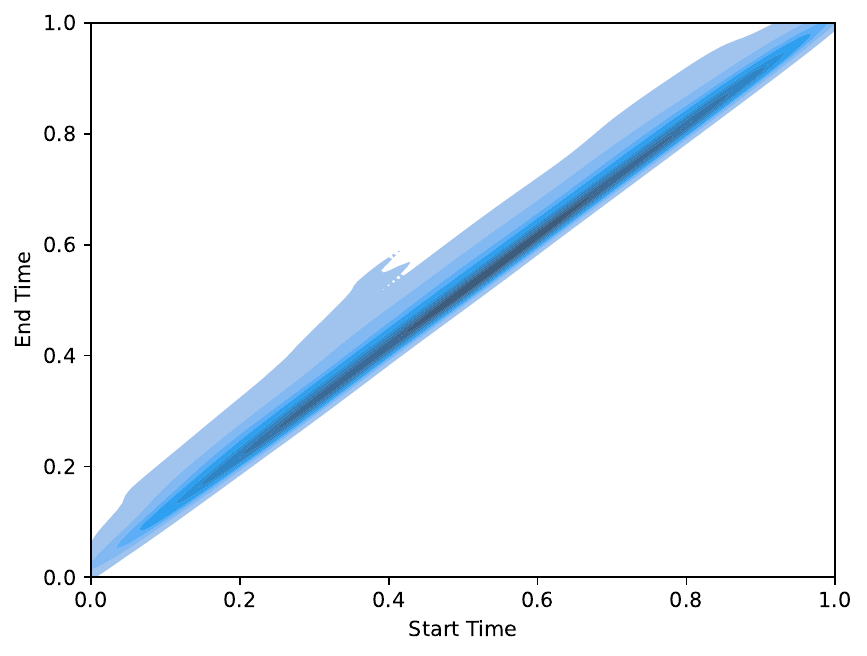}
    \caption{IAW Test}
    \label{fig:iaw_test}
  \end{subfigure}
  \caption{Kernel density estimate (KDE) plots of normalized start and end time distribution for different datasets, where the locations of the ground truth time span in YouCookII and IAW are more uniformly spread through the video compared with Charades STA, ActivityNet Caption and TACoS. Besides that, there are strong location priors illustrated as left bottom corner blob shown in (\subref{fig:charades_val},~\subref{fig:anet_val},~\subref{fig:tacos_val},~\subref{fig:tacos_test}). In contrast to the IAW and YouCookII, where the blob is actually spread through all the video sequence.}
  \label{fig:kde}
\end{figure}

\begin{table}[tb]
    \centering
    \scriptsize
    \newcolumntype{Z}{>{\centering\arraybackslash}X}
    \caption{Evaluation results across diverse datasets where each prediction spans the video's entire length.}
    \begin{NiceTabularX}{\linewidth}{
            @{} l | Z | *{3}{Z} | Z
        }
        \toprule
        \Block{2-1}{Dataset}
        &\Block{2-1}{Split}
        & \Block{1-3}{R@1, IoU=}
        &
        &
        & \Block{2-1}{mIoU}
        \\
        \cmidrule(lr){3-5}
        &
        & 0.3
        & 0.5
        & 0.7
        \\
        \midrule
        Charades-STA~\cite{gao2017tall}
        & Test
        & 27.10
        & 2.392
        & 0.027
        & 24.52
        \\
        ActivityNet-Caption~\cite{krishna2017dense} 
        & Val
        & 43.50
        & 23.20
        & 11.59
        & 32.29
        \\
        TACoS~\cite{rohrbach2014coherent} 
        & Val
        & 5.207
        & 1.195
        & 0.293
        & 8.491
        \\
        TACoS~\cite{rohrbach2014coherent} 
        & Test
        & 6.098
        & 1.600
        & 1.600
        & 8.759
        \\
        \midrule
        YouCook2~\cite{zhou2018towards} 
        & Val
        & 0.573
        & 0.057
        & 0.000
        & 6.437
        \\
        IAW~\cite{zhang2023aligning} 
        & Train
        & 0.365
        & 0.061
        & 0.010
        & 3.748
        \\
        IAW~\cite{zhang2023aligning} 
        & Val
        & 0.226
        & 0.045
        & 0.000
        & 3.640
        \\
        IAW~\cite{zhang2023aligning} 
        & Test
        & 0.199
        & 0.028
        & 0.000
        & 3.654
        \\
        \bottomrule
    \end{NiceTabularX}
    \label{tab:full_duration_prediction}
\end{table}

Yuan \etal~\cite{yuan2021closer} highlight that the conventional metric for TSGV tasks, $\text{R}@k,\text{IoU}=m$, contributes to unreliable benchmarking, primarily due to biases in dataset annotations as shown in~\cref{fig:kde}, such as the presence of excessively long ground-truth moments. Reflecting on this, our findings from predicting the full video duration as the predicted time span, as shown in~\cref{tab:full_duration_prediction}, demonstrate that this simplistic, non-learning approach yields comparatively high performance on datasets like Charades-STA~\cite{gao2017tall} and ActivityNet-Caption~\cite{krishna2017dense}. This suggests that first, the standard metric remains effective for the IAW and YouCook2 and second, that the IAW and YouCook2 pose a greater challenge due to a lack of such annotation bias.

\begin{table}[!htb]
    \centering
    \scriptsize
    \newcolumntype{Z}{>{\centering\arraybackslash}X}
    \caption{Evaluation results on ActivityNet Caption~\cite{krishna2017dense} \texttt{val\_2} split.}
    \begin{NiceTabularX}{\linewidth}{
            @{} l | *{3}{Z} | Z
        }
        \toprule
        \Block{2-1}{Method}
        & \Block{1-3}{R@1, IoU=}
        &
        &
        & \Block{2-1}{mIoU}
        \\
        \cmidrule(lr){2-4}
        & 0.3
        & 0.5
        & 0.7
        \\
        \midrule
        \Block{1-5}{Video Grounding Methods}
        \\
        \midrule
        MCN~\cite{anne2017localizing}
        & 39.35
        & 21.36
        & 6.43
        & 15.83
        \\
        CTRL~\cite{gao2017tall}
        & 47.43
        & 29.01
        & 10.34
        & 20.54
        \\
        TGN~\cite{chen2018temporally}
        & 43.81
        & 27.93
        & 11.86
        & 29.17
        \\
        QSPN~\cite{xu2019multilevel}
        & 52.13
        & 33.26
        & 13.43
        & -
        \\
        ABLR~\cite{yuan2019find}
        & 55.67
        & 36.79
        & -
        & 36.99
        \\
        DRN~\cite{zeng2020dense}
        & -
        & 45.45
        & 24.36
        & -
        \\
        2D-TAN~\cite{zhang2020learning}
        & 59.45
        & 44.51
        & 26.54
        & -
        \\
        VSLNet~\cite{zhang2020span}
        & 63.16
        & 43.22
        & 26.16
        & 43.19
        \\
        CPNet~\cite{li2021proposal}
        & -
        & 40.56
        & 21.63
        & 40.65
        \\
        BPNet~\cite{xiao2021boundary}
        & 58.98
        & 42.07
        & 24.69
        & 42.11
        \\
        CBLN~\cite{liu2021context}
        & 66.34
        & 48.12
        & 27.60
        & -
        \\
        TACI~\cite{shin2022learning}
        & -
        & 45.50
        & 27.23
        & -
        \\
        BMRN~\cite{seol2023bmrn}
        & 66.34
        & 48.47
        & 31.15
        & -
        \\
        \midrule
        \Block{1-5}{Dense Video Grounding Methods}
        \\
        \midrule
        BS~\cite{bao2021dense}
        & 62.53
        & 46.43
        & 27.12
        & -
        \\
        3D-TPN~\cite{bao2021dense}
        & 67.56
        & 51.49
        & 30.92
        & -
        \\
        DepNet~\cite{bao2021dense}
        & 72.81
        & 55.91
        & 33.46
        & -
        \\
        SVPTR~\cite{jiang2022semi}
        & 78.07
        & 61.70
        & 38.36
        & 55.91
        \\
        HSCNet~\cite{tan2023hierarchical}
        & \textbf{81.89}
        & \textbf{66.57 }
        & \textbf{44.03}
        & \textbf{69.71}
        \\
        PRVG~\cite{shi2024end}
        & 78.27
        & 61.15
        & 37.83
        & 55.62
        \\
        \midrule
        Ours
        & 79.00
        & 62.16
        & 37.60
        & 56.22
        \\
        \quad w/ Contrastive Alignment
        & 78.73
        & 62.53
        & 38.55
        & 56.39
        \\
        \bottomrule
    \end{NiceTabularX}
    \label{tab:activitynet}
\end{table}

\subsection{Results on ActivityNet-Captions}

Due to the lack of publicly available code for SVPTR~\cite{jiang2022semi}, HSCNet~\cite{tan2023hierarchical}, and PRVG~\cite{shi2024end} to test on IAW~\cite{zhang2023aligning} or YouCook2~\cite{zhou2018towards}, we instead opted to evaluate our model using the ActivityNet-Caption~\cite{krishna2017dense} dataset. The results, as shown in~\cref{tab:activitynet}, demonstrate that our model, although not specifically designed for the language grounding task (we adapted it by averaging word features), outperforms all referenced methods with the exception of HSCNet~\cite{tan2023hierarchical}, where the primary advantage of HSCNet lies in its multilevel hierarchical semantic alignment for visual-textual correspondence, which is not applicable for step diagrams in the IAW dataset.

\subsection{Self-Attention Visualization}

\begin{figure}[tb]
  \includegraphics[width=\linewidth]{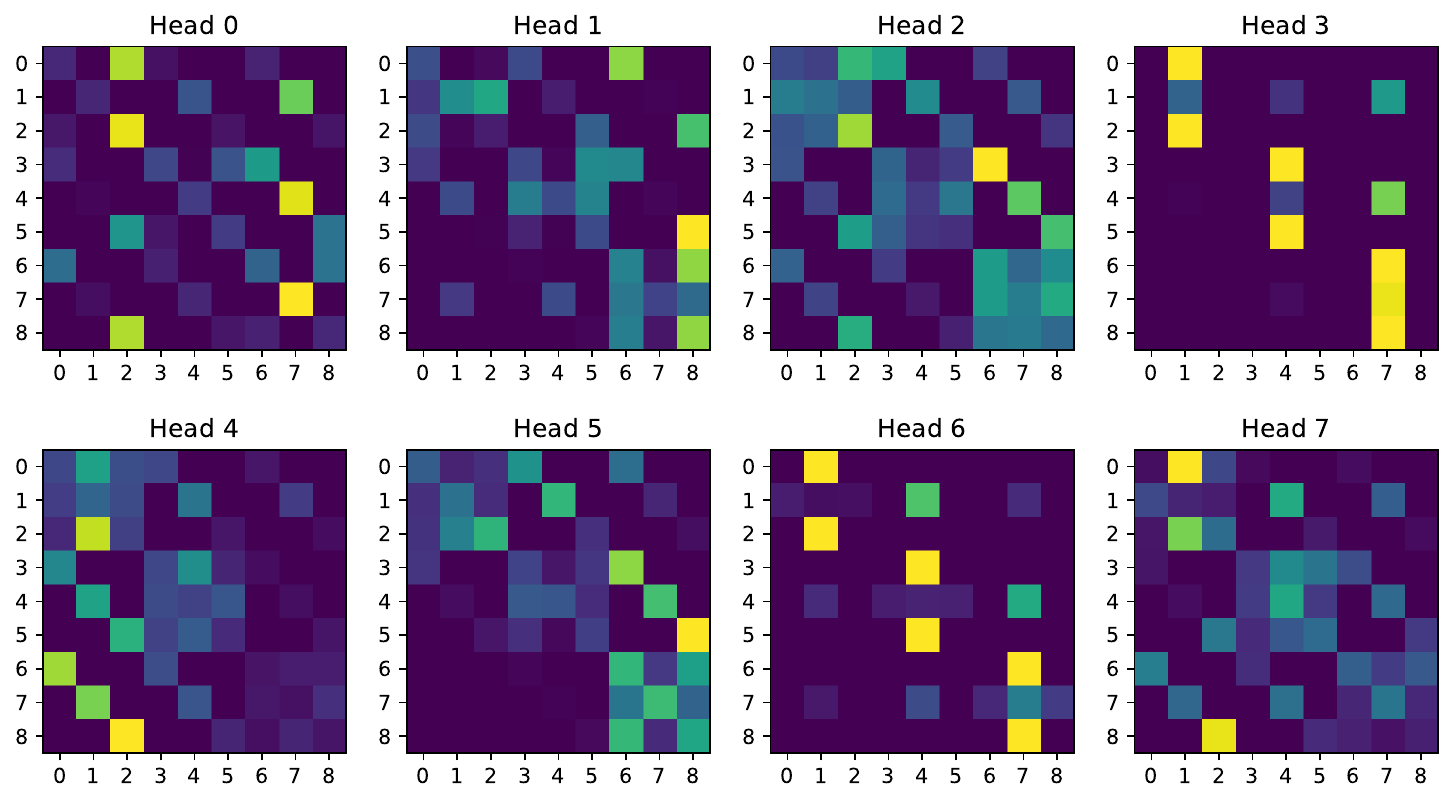}
  \caption{Visualization of last layer self-attention among composite queries for the same example shown in cross-attention visualization the main paper. The index on the axis denotes the corresponding composite query, \eg, 0 means composite query (1, 1) with diagram 1 and learnable query 1, 3 represents (2, 1), 7 denotes (3, 2) and so on. Composite queries 2, 4 and 7 get the highest scores at the end.}
\label{fig:sa}
\end{figure}

As shown in~\cref{fig:sa}, the hypothesised suppressive behaviors can be observed by the self-attention scores in head 3, 4 and 6. For example, in head 6, the highlighted self-attention scores between composite query 2 and other queries, indicating that composite query 2 gets the highest score at the end.

\begin{table}[ht]
    \centering
    \scriptsize
    \newcolumntype{Z}{>{\centering\arraybackslash}X}
    \caption{Ablation results for different types of $K$, $Q$ fusion on the IAW test split. \textit{add} and \textit{concate} denote for $K+Q$ and $K\oplus Q$ respectively. Notably, the concatenate is applied on head-level in the multi-head attention module.}
    \begin{NiceTabularX}{\linewidth}{
            @{} l | *{3}{Z} | Z
        }[code-before = {\rowcolor{gray!25}{4}}]
        \toprule
        \Block{2-1}{$K,Q$ Fusion Type}
        & \Block{1-3}{R@1, IoU=}
        &
        &
        & \Block{2-1}{mIoU}
        \\
        \cmidrule(lr){2-4}
        & 0.3
        & 0.5
        & 0.7
        \\
        \midrule
        add
        & \underline{35.73}
        & \underline{20.88}
        & \underline{8.141}
        & \underline{22.83}
        \\
        concat
        & \textbf{37.79}
        & \textbf{22.74}
        & \textbf{9.140}
        & \textbf{23.86}
        \\
        \bottomrule
    \end{NiceTabularX}
    \label{tab:attention_kq}
\end{table}

\begin{table}[ht]
    \centering
    \scriptsize
    \newcolumntype{Z}{>{\centering\arraybackslash}X}
    \caption{Ablation results for different number of decoder layers on the IAW test split.}
    \begin{NiceTabularX}{\linewidth}{
            @{} l | *{3}{Z} | Z
        }[code-before = {\rowcolor{gray!25}{7}}]
        \toprule
        \Block{2-1}{\# of Layers}
        & \Block{1-3}{R@1, IoU=}
        &
        &
        & \Block{2-1}{mIoU}
        \\
        \cmidrule(lr){2-4}
        & 0.3
        & 0.5
        & 0.7
        \\
        \midrule
        \quad 2
        & 31.62
        & 18.08
        & 6.370
        & 20.28
        \\
        \quad 3
        & 33.16
        & 18.28
        & 7.227
        & 21.26
        \\
        \quad 4
        & 34.48
        & 19.68
        & 7.741
        & 21.77
        \\
        \quad 5
        & 36.10
        & 20.54
        & 8.312
        & 22.82
        \\
        \quad 6
        & \underline{37.79}
        & \textbf{22.74}
        & \textbf{9.140}
        & 23.86
        \\
        \quad 7
        & \textbf{38.07}
        & \underline{22.05}
        & \underline{8.740}
        & \textbf{24.05}
        \\
        \quad 8
        & 37.76
        & 21.88
        & 8.683
        & \underline{23.91}
        \\
        \bottomrule
    \end{NiceTabularX}
    \label{tab:n_layers}
\end{table}

\begin{table}[ht]
    \centering
    \scriptsize
    \newcolumntype{Z}{>{\centering\arraybackslash}X}
    \caption{Ablation results for different number of learnable queries on the IAW test split.}
    \begin{NiceTabularX}{\linewidth}{
            @{} l | *{3}{Z} | Z
        }[code-before = {\rowcolor{gray!25}{3}}]
        \toprule
        \Block{2-1}{\# of Queries}
        & \Block{1-3}{R@1, IoU=}
        &
        &
        & \Block{2-1}{mIoU}
        \\
        \cmidrule(lr){2-4}
        & 0.3
        & 0.5
        & 0.7
        \\
        \midrule
        \quad 3
        & \underline{37.79}
        & \textbf{22.74}
        & \textbf{9.140}
        & \underline{23.86}
        \\
        \quad 5
        & 36.85
        & 21.91
        & \textbf{9.140}
        & 23.82
        \\
        \quad 10
        & \textbf{37.82}
        & 21.19
        & 8.655
        & \textbf{24.00}
        \\
        \quad 15
        & 37.13
        & \underline{22.51}
        & \underline{8.883}
        & 23.52
        \\
        \quad 20
        & 36.42
        & 22.11
        & 8.826
        & 23.45
        \\
        \quad 30
        & 36.82
        & 21.71
        & 8.598
        & 23.45
        \\
        \bottomrule
    \end{NiceTabularX}
    \label{tab:n_queries}
\end{table}

\subsection{Ablation Study}

We also evaluate the performance with different key value fusing type for both self-attention and cross-attention, as shown in~\cref{tab:attention_kq}, where the concatenate then project them back operation outperforms the simple addition. Additionally, we show the results for different numbers of decoder layers and learnable queries  in~\cref{tab:n_layers,tab:n_queries}, respectively.

\subsection{Examples}

\begin{figure*}[tb]
  \centering
  \begin{subfigure}{0.8\linewidth}
    \includegraphics[width=\linewidth]{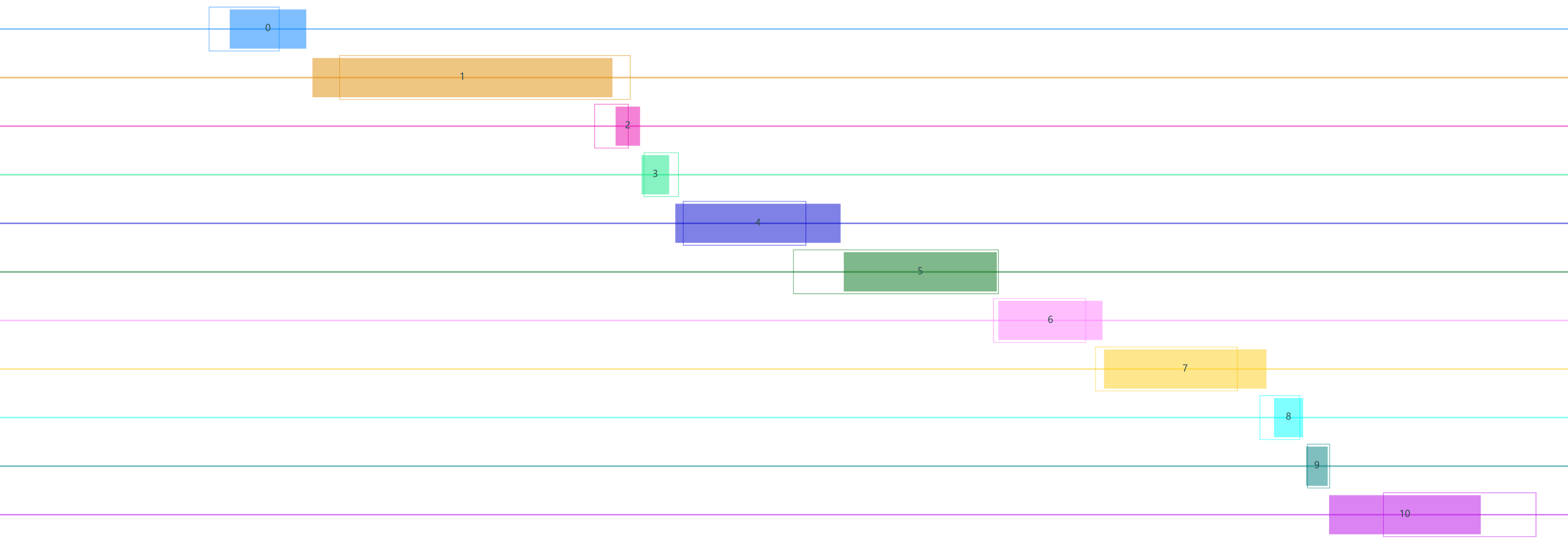}
    \caption{A successful grounding example between video \href{https://www.youtube.com/watch?v=TEdm3OyeWjY}{TEdm3OyeWjY} and furniture \href{https://www.ikea.com/au/en/p/teodores-chair-white-70350938/}{70350938}.}
    \label{fig:good-1}
  \end{subfigure}
  \\
  \begin{subfigure}{0.8\linewidth}
    \includegraphics[width=\linewidth]{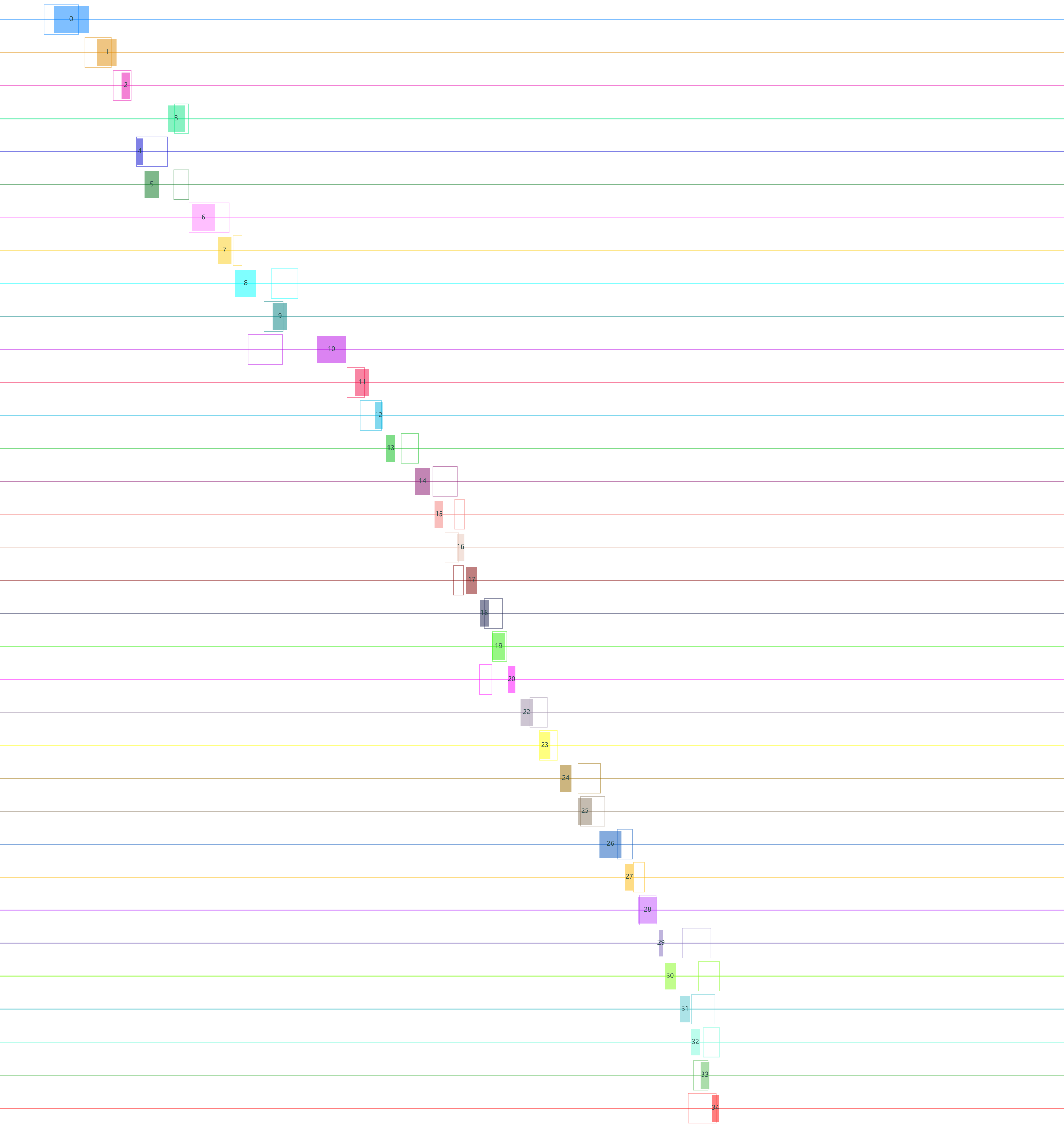}
    \caption{A successful grounding example between video \href{https://www.youtube.com/watch?v=nPtQ28088zw}{nPtQ28088zw} and furniture \href{https://www.ikea.com/au/en/p/hemnes-chest-of-8-drawers-white-stain-80355695/}{80355695}.}
    \label{fig:good-2}
  \end{subfigure}
  \caption{Qualitative result of two successful examples. The horizontal axis represents the timeline of the video. Each row corresponds to a step diagram where the solid rectangle denotes the ground truth and top-1 time span predictions are represented by bounding boxes with the same color.}
  \label{fig:good-example}
\end{figure*}

\begin{figure*}[tb]
  \centering
  \begin{subfigure}{0.8\linewidth}
    \includegraphics[width=\linewidth]{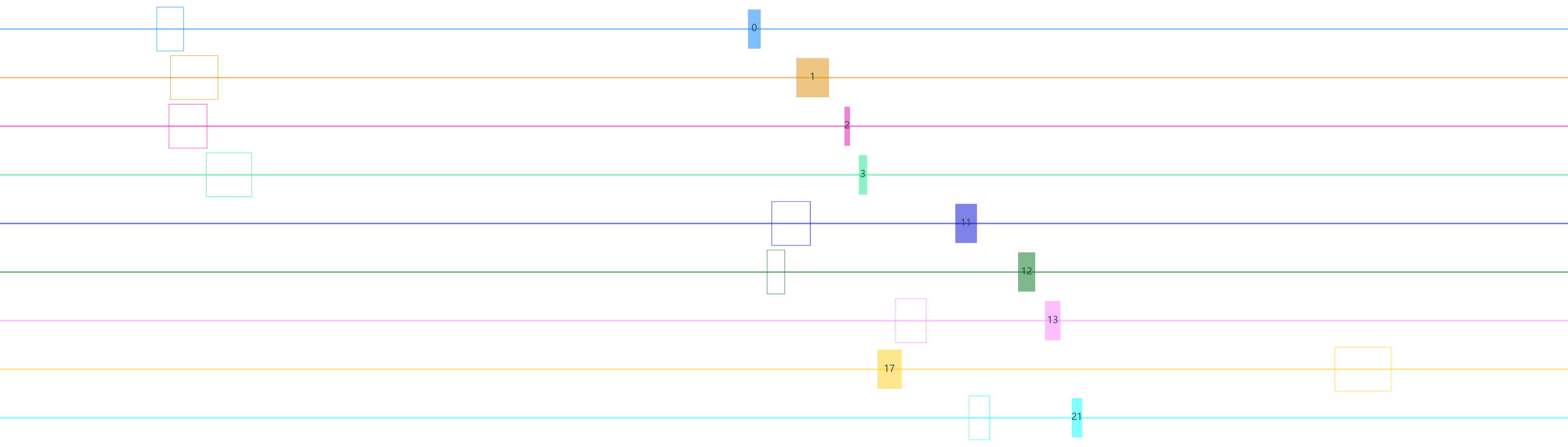}
    \caption{A failed grounding example between video \href{https://www.youtube.com/watch?v=jX8jSx9wg-8}{jX8jSx9wg-8} and furniture \href{https://www.ikea.com/au/en/p/fjaellbo-shelving-unit-black-80339295/}{80339295}.}
    \label{fig:bad-1}
  \end{subfigure}
  \\
  \begin{subfigure}{0.8\linewidth}
    \includegraphics[width=\linewidth]{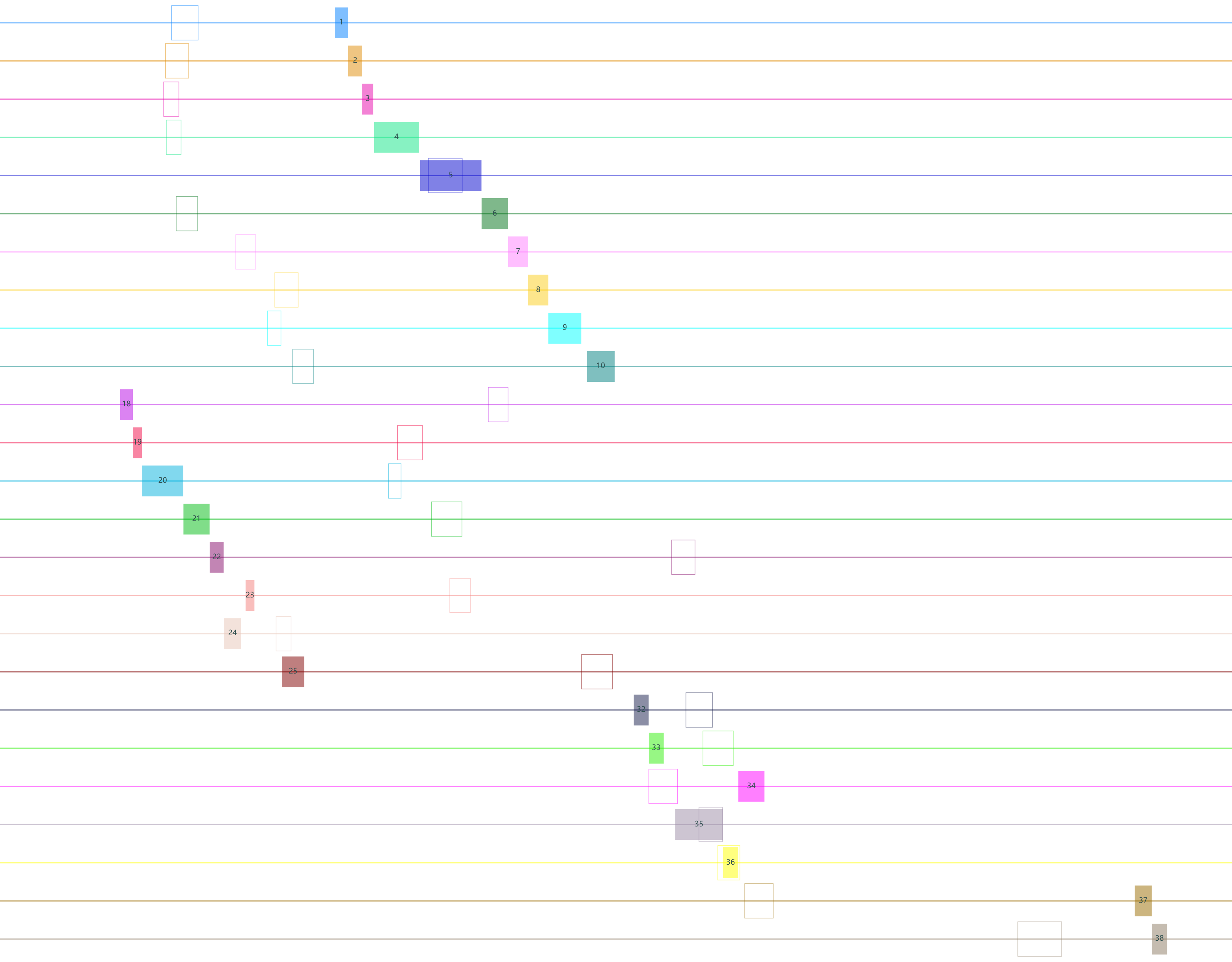}
    \caption{A failed grounding example between video \href{https://www.youtube.com/watch?v=rqrS7n45oA0}{rqrS7n45oA0} and furniture \href{https://www.ikea.com/au/en/p/besta-storage-combination-with-doors-white-lappviken-stubbarp-white-s49187459/}{s49187459}.}
    \label{fig:bad-2}
  \end{subfigure}
  \caption{Qualitative result of two failed examples. The horizontal axis represents the timeline of the video. Each row corresponds to a step diagram where the solid rectangle denotes the ground truth, and the top-1 time span predictions are represented by bounding boxes with the same color.}
  \label{fig:bad-example}
\end{figure*}

We show another two successful and failed examples in~\cref{fig:good-example,fig:bad-example}, respectively.

{
\small
\bibliographystyle{ieee_fullname}
\bibliography{supplementary}
}